\documentclass[runningheads]{llncs}

\usepackage{eccv}

\usepackage{multirow}
\usepackage[export]{adjustbox}
\usepackage[dvipsnames]{xcolor}
\usepackage{indentfirst}
\usepackage{colortbl}
\usepackage{subcaption}

\definecolor{darkpink}{rgb}{0.91, 0.33, 0.5}
\definecolor{darkpastelgreen}{rgb}{0.01, 0.75, 0.24}

\usepackage{eccvabbrv}

\usepackage{graphicx}
\usepackage{booktabs}

\usepackage[accsupp]{axessibility}  

\usepackage[pagebackref,breaklinks,colorlinks,citecolor=eccvblue]{hyperref}

\usepackage{orcidlink}

\begin{document}

\title{ASteISR: Adapting Single Image Super-resolution Pre-trained Model for Efficient Stereo Image Super-resolution} 

\titlerunning{ASteISR}

\author{Yuanbo Zhou \inst{1} \and
Yuyang Xue \inst{2} \and
Wei Deng\inst{3} \and Xinlin Zhang \inst{1}  \and  Qinquan Gao\inst{1,3} \and Tong Tong\inst{1,3} \thanks{Corresponding author. T. Tong (ttraveltong@gmail.com), Y. Zhou (webbozhou@gmail.com),}}

\authorrunning{Y. Zhou Author et al.}

\institute{Fuzhou University \and
University of Edinburgh \and
Imperial Vision Technology}

\maketitle

\begin{abstract}
 Despite advances in the paradigm of pre-training then fine-tuning in low-level vision tasks, significant challenges persist particularly regarding the increased size of pre-trained models such as memory usage and training time. Another concern often encountered is the unsatisfying results yielded when directly applying pre-trained single-image models to multi-image domain. In this paper, we propose a efficient method for transferring a pre-trained single-image super-resolution (SISR) transformer network to the domain of stereo image super-resolution (SteISR) through a parameter-efficient fine-tuning (PEFT) method. Specifically, we introduce the concept of stereo adapters and spatial adapters which are incorporated into the pre-trained SISR transformer network. Subsequently, the pre-trained SISR model is frozen, enabling us to fine-tune the adapters using stereo datasets along. By adopting this training method, we enhance the ability of the SISR model to accurately infer stereo images by 0.79dB on the Flickr1024 dataset. This method allows us to train only $4.8\%$ of the original model parameters, achieving state-of-the-art performance on four commonly used SteISR benchmarks. Compared to the more complicated full fine-tuning approach, our method reduces training time and memory consumption by $57\%$ and $15\%$, respectively.
  \keywords{Parameter-efficient fine-tuning \and Stereo image super-resolution \and Transfer learning}
\end{abstract}

\section{Introduction}
\label{sec:intro}

   \begin{figure}[ht]
    \centering
    \includegraphics [width=0.82\textwidth]{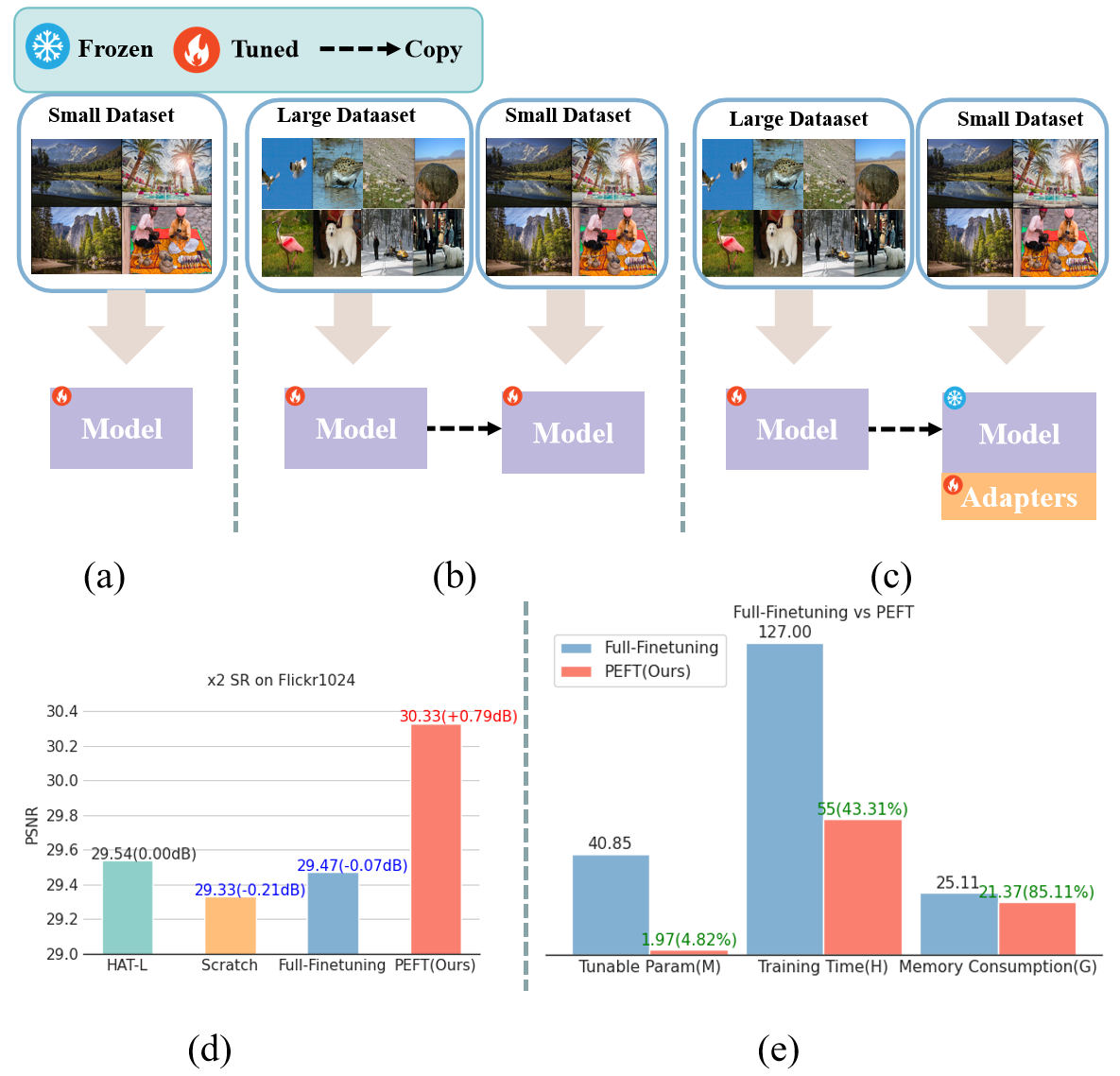}
    \caption{Common training strategies in low-level vision include: (a) training from scratch, (b) full fine-tuning (pre-training then fine-tuning), and (c) parameter-efficient fine-tuning (Ours). (d) X2 results of SteISR on Flickr1024 dataset; (e) comparison of the tunable parameters, training time and memory.}
    \label{fig:pretrain_vs_peft}
    \end{figure}
The ``pre-training then fine-tuning'' pipeline has been consistently playing a crucial role in both the field of Natural Language Processing (NLP) and high-level computer vision (CV). By fine-tuning pre-trained models using limited downstream data, one can effectively improve the model's accuracy in downstream tasks while mitigating overfitting. Prominent examples of this approach include language models like BERT~\cite{devlin2018bert} and GPT~\cite{radford2019language}, as well as visual models such as Moco~\cite{chenempirical}, SimCLR~\cite{chen2020big}, and DINO~\cite{oquab2023dinov2}. However, the growing size of these models poses challenges regarding GPU resources necessary for full fine-tuning.

Recently, NLP researchers started researching on parameter-efficient fine-tuning (PEFT) techniques as a solution to mitigate the complexities of fine-tuning large models. There are notable contributions in this area include adapter-like methods~\cite{houlsby2019parameter, pfeiffer2020adapterfusion, ruckle2020adapterdrop} and the LoRA series~\cite{hu2021lora, zhang2023adaptive, dettmers2023qlora}. The fields of high-level vision research have noted promising progress in PEFT, including VPT~\cite{jia2022visual}, AIM~\cite{yang2023aim}, and CoOp~\cite{zhou2022learning}. This instrument significantly reduces the computational and memory resources necessary to fine-tune large models. In addition, these PEFT methods provide competitive performance, even outperforming fully fine-tuned models in some instances. As such, these advancements are beginning to reshape the landscape of model fine-tuning, offering more efficient and advantageous strategies.

However, in the low-level domain, there have been limited investigations into PEFT methods. The common practice in this area predominantly involves either training models from scratch on fixed datasets or reverting to the pre-training then fine-tuning model~\cite{li2021efficient, chen2021pre, chen2023activating}, as shown in~\cref{fig:pretrain_vs_peft}. The domain of SteISR, a subfield harnessing potential applications in sectors such as autonomous driving~\cite{chuah2022semantic} and robotics~\cite{cosner2022self}, it has attracted increasing attention from researchers, resulting in the development of several influential methods~\cite{chu2022nafssr, cheng2023hybrid, zhou2023stereo, jin2022swinipassr, dai2021feedback, lin2023steformer}. It is noteworthy that almost all of these methods start training from scratch on fixed datasets such as Flickr1024~\cite{wang2019flickr1024} with 800 pairs and Middlebury~\cite{scharstein2014high} with 60 pairs. Regrettably, due to the smaller scale and low-quality training data for SteISR, the resulting models in this field underperform in texture recovery compared to SISR methods, significantly impeding the progress of SteISR. In addition, the SteISR models are also growing increasingly large, making the training costs challenging to manage~\cite{cheng2023hybrid}. As such, it is evident that there is a pressing need for research and development in effective and efficient learning methods tailored for this domain.

To address the dilemma of SteISR, this study initially attempts to fully fine-tune the pre-trained SISR network HAT~\cite{chen2023activating}, which is a SISR model pre-trained on large dataset ImageNet~\cite{deng2009imagenet}, on the SteISR datasets Flickr1024~\cite{wang2019flickr1024} and Middlebury~\cite{scharstein2014high} to evaluate the effectiveness of the pre-training then fine-tuning method. However, the fine-tuned model exhibits inferior performance compared to the original one, as shown in~\cref{fig:pretrain_vs_peft}(d). This performance gap leads an important question: does the pre-training then fine-tuning method not work for low-level vision or specifically for SteISR? To explore the reasons behind this, the HAT model~\cite{chen2023activating} is trained from scratch on the same SteISR datasets, and through comparative tests, it is found that the performance of the model trained from scratch is significantly lower than that of the fine-tuned model, as shown in~\cref{fig:pretrain_vs_peft}(d). Thus, it can be concluded that the pre-training then fine-tuning method is not the cause. On the contrary, the low-quality stereo image datasets play a pivotal role. This phenomenon bears some similarity to the concept of  ``catastrophic forgetting".

Inspired by the PEFT method utilized in the field of NLP, this study proposes a method involving the introduction of a small number of additional parameters into the pre-trained SISR model. Only these introduced parameters are then updated during the fine-tuning process. Specifically, stereo adapters and spatial adapters are introduced into the frozen SISR transformer network HAT~\cite{chen2023activating} and fine-tuned using stereo image datasets. This enables the SISR model to effectively handle stereo images. As the original model's parameters remain unchanged, a substantial portion of the knowledge from the original SISR model is preserved, thereby mitigating the issue of ``catastrophic forgetting". Furthermore, by training solely the introduced additional parameters, memory consumption and training time are significantly reduced, resulting in enhanced training efficiency. Leveraging this methodological advancement, only \textbf{1.97M} parameters are trained in this study, accounting for approximately \textbf{3\%} of the parameters utilized by the previous state-of-the-art model, HTCAN~\cite{cheng2023hybrid}. Our proposed approach achieves outstanding performance across all benchmarks. In summary, the main contributions of this study are as follows:

\begin{itemize}
 
  \item We propose a PEFT method, adapting a single-image model to a multi-image model, which may offer an insight for tasks such as video SR, video denoising, and video deblurring in large model era.
 
  \item A method is developed that enhances the capability of SISR model to effectively process stereo images. This method transfers successful SISR models to the domain of SteISR, thereby facilitating the simultaneous progress in both stereo and single-image super-resolution.
 
  \item Compared to HTCAN~\cite{cheng2023hybrid}, our proposed method require training the model parameters by approximately $3\%$ and the $25\%$ of the GPU resources while achieving comparable performance. 
  \item Our method achieves the state-of-the-art performance in SteISR.
\end{itemize}

\section{Related Works}
\label{sec:relate_works}
\subsection{Stereo Image Super-resolution(SteISR)}

The progression of SteISR has been marked by several junctures. StereoSR~\cite{jeon2018enhancing} reached a milestone by incorporating horizontal pixel shifting of images and feeding the manipulated images into a reconstruction network. Wang et al.~\cite{wang2019flickr1024} then contributed by presenting Flickr1024, a large-scale dataset that facilitated further advancements. Subsequently, researchers shifted their focus towards the development of stereo attention modules. Noteworthy works in this area include the integration of parallax attention mechanism by Wang et al.~\cite{wang2019learning}, the bidirectional parallax attention module proposed by Wang et al.~\cite{wang2021symmetric}, and the stereo cross-attention module introduced by Chu et al.~\cite{chu2022nafssr}. Furthermore, Zhou et al.~\cite{zhou2023stereo} improved the performance of SteISR models with the stereo cross-global attention module. Building upon the success of Transformers, researchers gradually incorporated Transformers into the realm of SteISR. For instance, Jin et al.~\cite{jin2022swinipassr} proposed SwinIPASSR, which leverages Transformers to enhance global perception. Later, Cheng et al.~\cite{cheng2023hybrid} presented a two-stage hybrid network that combines Transformers with CNNs, achieving state-of-the-art performance through cascaded output.

Although the previous works have been achieved in the field of SteISR, they ignored the importance of pre-training models. The pipeline of pre-training then fine-tuning has gained popularity in the field of NLP due to the emergence of innovative developments such as Transformer~\cite{vaswani2017attention} and effective self-supervised training strategies like BERT~\cite{devlin2018bert} and GPT~\cite{radford2019language}. Gradually extended into high-level computer vision applications. Recent studies~\cite{dosovitskiy2020image, liu2021swin} have explored the use of Transformers trained on large-scale image classification datasets, combined with a small amount of task-specific data for efficient transfer learning. These methods have shown promising effects in various high-level computer vision tasks, including medical image classification~\cite{wang2021boundary}, segmentation~\cite{wang2021boundary}, and detection~\cite{mathai2022lymph}. Moreover, the introduction of the CLIP~\cite{radford2021learning} model has spurred the evolution of the pre-training and fine-tuning pipeline towards the multi-modal direction.

\subsection{Parameter-efficient Fine-tuning (PEFT)}

To improve the efficiency of fine-tuning method, many PEFT methods have emerged in the field of NLP. These PEFT methods can be categorized into three main categories: reparameterization, selectively partial parameters updating, and extra parameters addition methods. Reparameterization methods, exemplified by Hu et al.~\cite{hu2021lora}, proposed a method named LoRA that employs low-rank decomposition to represent parameter changes in the model, enabling indirect training of large models with minimal parameters. Subsequently, Zhang et al.~\cite{zhang2023adaptive} introduced AdaLoRA, which dynamically adjusts the rank of a matrix to enhance performance of the models without increasing training complexity. Addressing the complexity of model training, Tim et al.~\cite{dettmers2023qlora} further compressed it utilizing constant re-quantization. Selective parameter update methods include BitFit~\cite{zaken2021bitfit} and CSFT~\cite{ansell2021composable}, which minimize the cost of fine-tuning extensive models by focusing solely on biases. Another method involves adding additional parameters solely for updates, as demonstrated by Neil et al. ~\cite{houlsby2019parameter} proposing the Adapter tuning, which included two adapter layers in the Transformer and trains only those layers. For integrating knowledge from multiple tasks, Jonas et al. ~\cite{pfeiffer2020adapterfusion} proposed the addition of an Adapter Fusion layer after multiple adapter layers. Later, to improve computational efficiency, Andreas et al. ~\cite{ruckle2020adapterdrop} introduced AdapterDrop.

The concept of leveraging breakthroughs in the field of NLP has gradually extended to high-level computer vision as well. One notable work in this regard is CoOp~\cite{zhou2022learning}, wherein the authors fine-tuned the CLIP multimodal pre-trained model using prompts. This approach effectively enhanced the model's classification performance in various downstream tasks. Another work by Jia et al.~\cite{jia2022visual} introduced learnable prompt tokens into the input embedding vectors and selectively updated the weights of the prompt token and the final classification head. This method significantly improved the processing capability of downstream tasks. Contrasting with prior works, Yang et al.~\cite{yang2023aim} proposed AIM, which facilitates video action recognition by incorporating pre-trained image models with additional components such as spatial adapters, temporal adapters, and joint adapters, achieving competitive performance.

However, there is a relative scarcity of work on PEFT in the low-level vision field. Therefore, this paper mainly focuses on the field of SteISR as the starting point, with the aim of exploring PEFT methods specifically tailored for low-level vision.


\section{Proposed Method}

 \begin{figure*}[ht]
    \centering
    \includegraphics [width=0.98\textwidth]{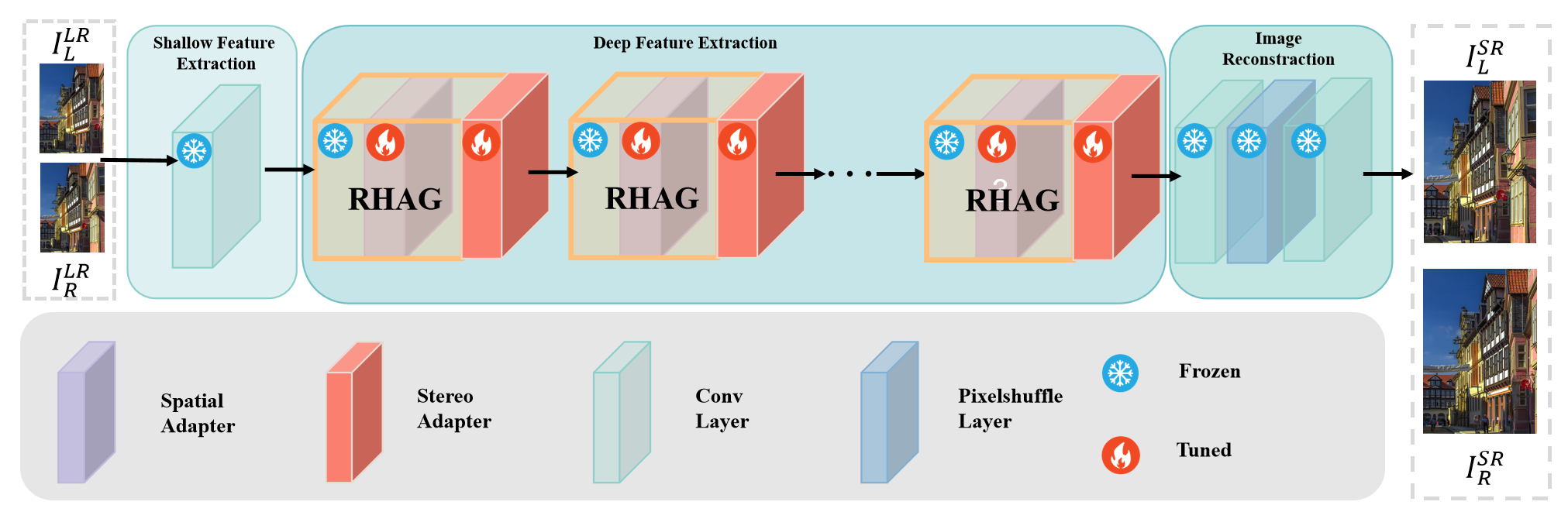}
    \caption{The ASteISR framework, achieving SteISR by adding spatial and stereo adapters to the pre-trained SISR HAT model~\cite{chen2023activating}.}
    \label{fig_framework}
    \end{figure*}

\noindent \textbf{The main idea} is to effectively adapt the pre-trained SISR model to the SteISR domain. To counter the problem of single-image model can not accurately infer stereo images, we introduce the stereo adapters. Besides, we integrate the spatial adapter to further improve the performance of SteISR.

\subsection{Method Overview}

To effectively adapt SISR models to the SteISR domain, this paper introduces two essential modules: the stereo adapter and the spatial adapter. The specific positions for their insertion can be observed in~\cref{fig_framework}. From the illustration, it is evident that the stereo adapter is integrated after the Residual Hybrid Attention Groups (RHAG) ~\cite{chen2023activating} module (shown in red), while the spatial adapter is embedded within the RHAG module (shown in purple). The stereo adapter's role is to facilitate the joint fusion of knowledge from both the left and right viewpoints, enabling efficient processing of stereo images by the SISR model. The spatial adapter further refines the information fused by the stereo adapter. Detailed explanations of these two components will be presented in the following sections.

\subsection{Differences between SISR and SteISR}
\captionsetup[subfigure]{labelformat=empty}
\label{sec:dsisrstesr}
        \begin{figure}
          \centering
              \begin{subfigure}[b]{0.49\textwidth}
                \includegraphics[width=\textwidth]{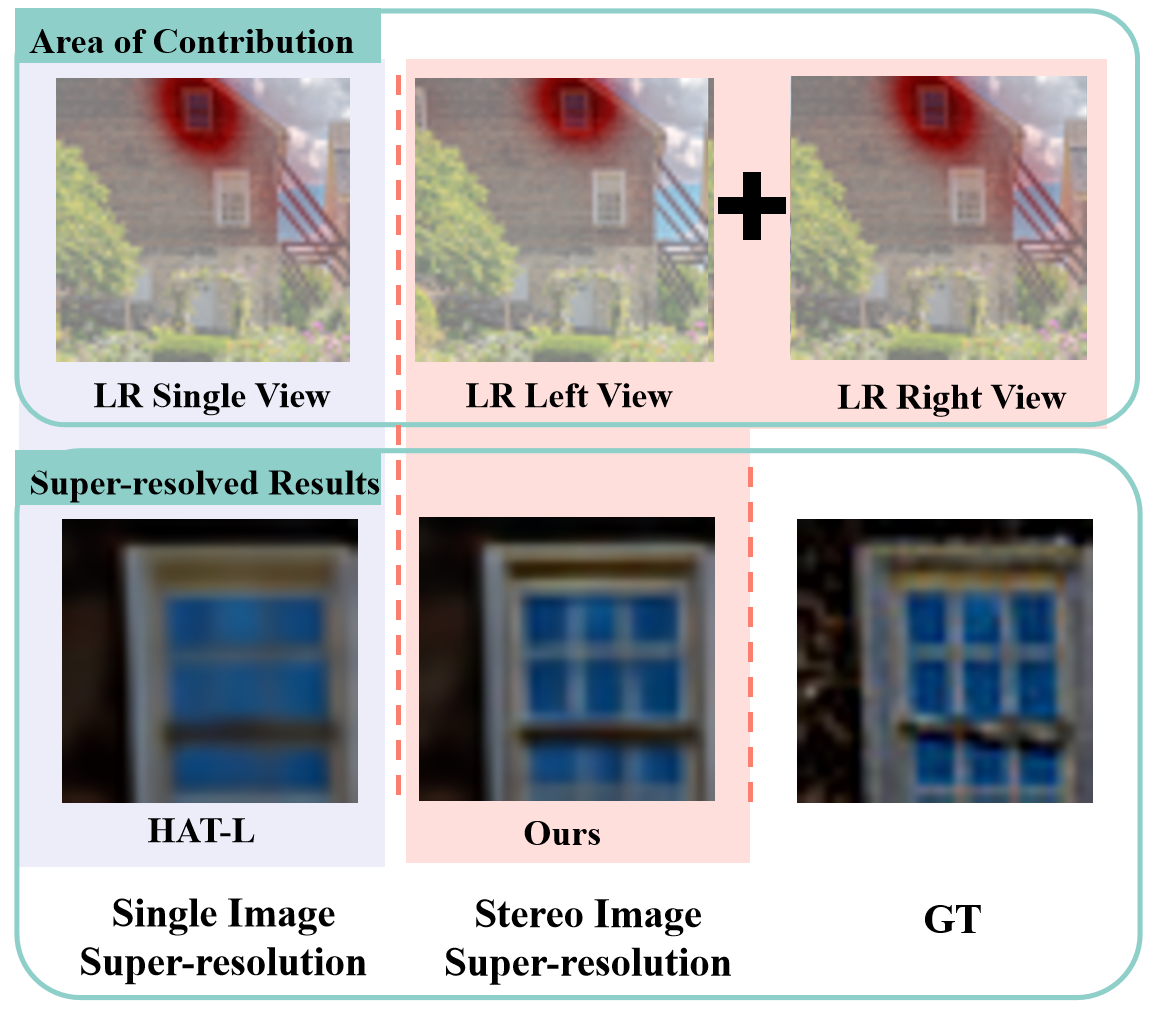}
                \subcaption{(a) The difference between SteISR and SISR}
                \label{fig:sisrvssteisr-fig:stereo_adapter:sub1}
              \end{subfigure}
          \hfill
              \begin{subfigure}[b]{0.49\textwidth}
                \includegraphics[width=\textwidth]{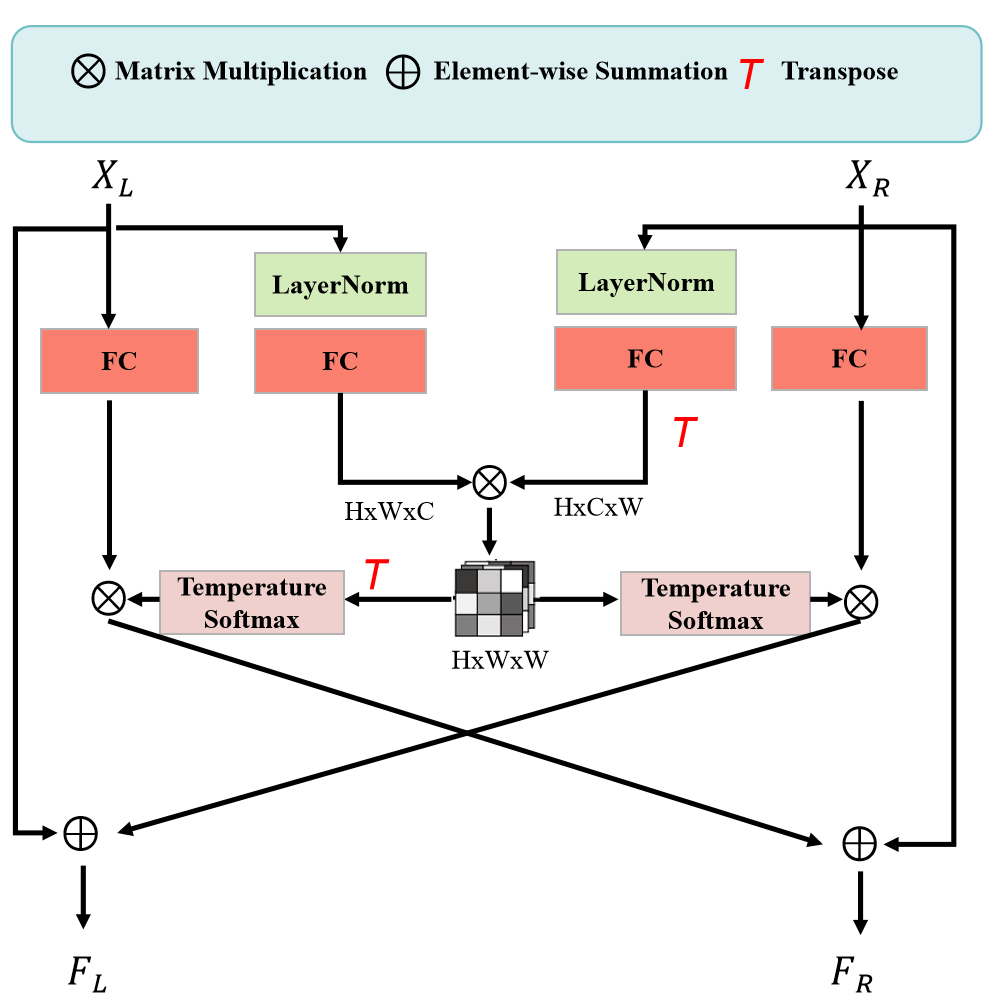}
                 \subcaption{(b) The detail of stereo adapter}
                \label{fig:sisrvssteisr-fig:stereo_adapter:sub2}
              \end{subfigure}
          \caption{The principle idea of using stereo adapter. (a) Single image super-resolution (SISR) and stereo image super-resolution (SteISR) differ in their utilization of knowledge for restoration and reconstruction. SISR relies solely on intra-image knowledge, while SteISR leverages both intra-image and cross-image knowledge. (b) The stereo adapter, which is to integrate information from the left and right views.}
   \label{fig:sisrvssteisr-fig:stereo_adapter}
   \end{figure}

The comparative analysis of SISR and SteISR lies primarily hinges on the utilization of input data by the models. SISR's super-resolution quality depends entirely on the single input image, which confines the model's capacity for texture reconstruction to its intra-image knowledge. Conversely, SteISR leverages both intra-image and inter-image knowledge, enabling the recapture of diminished textures. Consequently, in terms of texture restoration, SteISR should not be theoretically inferior to SISR. To demonstrate this point, we performed a visualization attribution analysis, examining SISR model HAT~\cite{chen2023activating} and the SteISR using Local Attention Map (LAM) ~\cite{gu2021interpreting}. \cref{fig:sisrvssteisr-fig:stereo_adapter} illustrates that SteISR surpasses SISR in terms of texture restoration ability. The region maps reveal the SteISR model's accurate identifications at discerning the requisite intra-image and inter-image knowledge, underscoring the significance of this dual knowledge integration for SteISR.

 \subsection{Stereo Adapter}

As outlined in~\cref{sec:dsisrstesr}, the key distinction SISR and SteISR is SteISR's ability to aggregate inter-image knowledge. Therefore, augmenting the SISR model with the ability to collect stereo information becomes imperative. Since a pre-trained SISR model lacks the capacity to capture inter-image information, this paper introduces a stereo adapter that can effectively capture and incorporate inter-image knowledge. By integrating this adapter after the RHAG module~\cite{chen2023activating}, the accumulation of stereo knowledge is accomplished. The stereo adapter primarily comprises of LayerNorm (LN)~\cite{ba2016layer}, fully connected (FC) layers, and Softmax with a temperature coefficient~\cite{zhou2023stereo}, as shown in ~\cref{fig:sisrvssteisr-fig:stereo_adapter}. The specific computation process can be represented by~\cref{equ:steadapter}.

    \begin{equation}
    \begin{aligned}
    & F_{R \rightarrow L}=\operatorname{TA}\left( W_1^{L} \overline{X}_{L}, W_1^{R} \overline{X}_{R}, W_2^{R} X_{R}\right), \\
    & F_{L \rightarrow R}=\operatorname{TA}\left( W_1^{R} \overline{X}_{R}, W_1^{L} \overline{X}_{L}, W_2^{L} X_{L}\right), \\
    & F_{L} = {\gamma}_{L} F_{R \rightarrow L} +  X_{L}, \\
    & F_{R} = {\gamma}_{R} F_{L \rightarrow R} +  X_{R}, \\
    \label{equ:steadapter}
    \end{aligned}
    \end{equation}

    \noindent where  $\overline{X}_{L} = LN(X_{L})$, $\overline{X}_{R} = LN(X_{R})$. $W_1^{L}$, $W_1^{R}$, $W_2^{L}$ and $W_2^{R}$ are projection matrices. ${\gamma}_{L}$ and ${\gamma}_{R}$ represent learnable scaling parameters. $\operatorname{TA}$ refers to the temperature attention module, which can be represented by \cref{equ:TA}.

    \begin{equation}
    \operatorname{TA}(Q, K, V)=\operatorname{softmax}\left( \tau Q K^T / \sqrt{C} \right) V,
    \label{equ:TA}
    \end{equation}

    \noindent where $\tau$ is a hyperparameter representing the temperature coefficient. This addition serves to augment the entropy of the fused features post-softmax, akin to the idea of knowledge distillation. 

\subsection{Spatial Adapter}
    \begin{figure}[h]
    \centering
    \includegraphics [width=0.75\textwidth]{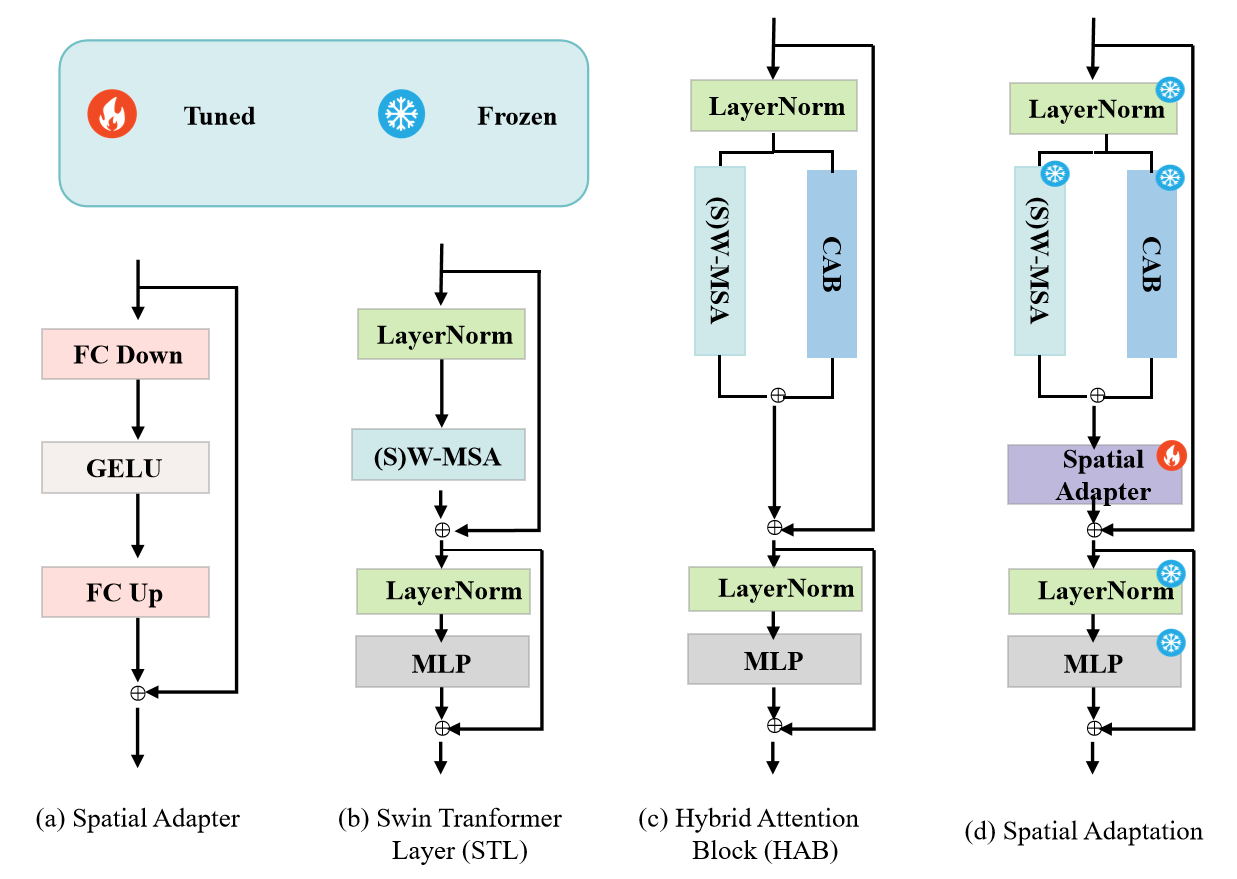}
    \caption{By incorporating a spatial adapter (a) into the HAB (c), the SISR model can efficiently handle stereo images. The main distinction between HAB and the STL arises from the inclusion of a Channel Attention Block (CAB) in the HAB.}
    \label{fig:spatial_adapter}
    \end{figure}

In theory, the HAT model, pre-trained on ImageNet~\cite{deng2009imagenet} and DF2K~\cite{lim2017enhanced, agustsson2017ntire}, already possesses the ability to handle single low-resolution images. The only requirement is to equip it with the capability to process stereo images. In other words, adding the stereo adapter would be sufficient, eliminating the need for spatial adaptation. However, our experiments have demonstrated that incorporating spatial adapter can further refine the information fused by the stereo adapter, thus enhancing the model's performance in processing stereo images.

To illustrate the process of spatial adapter, ~\cref{fig:spatial_adapter} provides a clear demonstration. We shall commence with the Swin Transformer Layer (STL)~\cite{liu2021swin}. Given a low-resolution image $I_{LR}\in \mathbf{R}^{hwc} $, it is divided into non-overlapping patches and undergoes patch embedding, resulting in $I_{p}\in \mathbf{R}^{ND}$, where $N$ represents the width multiplied by the height of the small patch, and $D$ represents the dimension of the embedding. Subsequently, $I_{p}$ is utilized as the input for the STL, which involves consecutive steps of LN, Window Multi-Headed Self Attention (W-MSA) / Shifted Window Multi-Headed Self Attention (SW-MSA), and Multilayer Perceptron (MLP), resulting in  $I_{p}^{\prime}$ with identical dimensionality to the initial input. Typically, W-MSA and SW-MSA are paired together, and the entire process can be represented by~\cref{equ:stl}.

\begin{equation}\label{equ:stl}
\begin{aligned}
    &\hat{I_{p}} = \operatorname{Attention}(\operatorname{LN(I_{p})})+I_{p},  \\
    &I_{p}^{\prime} = \operatorname{MLP}(\operatorname{LN}(\hat{I_{p}}))+\hat{I_{p}}, \\
\end{aligned}
\end{equation}

\noindent where Attention refers SW-MSA/W-MSA.

Similar to the approach used in HAT~\cite{chen2023activating}, we have incorporated a channel attention block (CAB) in parallel with S(W)-MSA at both ends. \cref{fig:spatial_adapter}(c) illustrates this configuration. The CAB consists of two convolutional layers with GELU activation function ~\cite{hendrycks2016gaussian}  and a channel attention module (CAM)~\cite{chen2023activating}. This upgraded module is referred to as the hybrid attention block (HAB), which enhances the capability of SISR by replacing the STL. Consequently, by introducing a spatial adapter based on the HAB, we further refine the information fused by the stereo adapter.

The spatial adapter is composed of two FC layers and employs the GELU activation function, as shown in~\cref{fig:spatial_adapter}(a). The first FC layer compresses the input dimensions, while the second FC layer restores the compressed tensor. Unlike the typical approach in NLP studies~\cite{houlsby2019parameter, pfeiffer2020adapterfusion}, where adapters are typically inserted after the Attention module and the MLP of the Transformer, in this study, we observed that inserting the adapter after the MLP has minimal impact on the final results. As a result, we only insert the spatial adapter after the hybrid attention (HA), as shown in \cref{fig:spatial_adapter}. The specific calculation process can be represented by ~\cref{equ:stlwa}.

\begin{equation}\label{equ:stlwa}
\begin{aligned}
    &\hat{I_{p}} = \operatorname{Spaital-Adapter}{(\operatorname{HA}(\operatorname{LN(I_{p})}))}+I_{p},  \\
    &I_{p}^{\prime} = \operatorname{MLP}(\operatorname{LN}(\hat{I_{p}}))+\hat{I_{p}}, \\
\end{aligned}
\end{equation}

\noindent where HA can be expressed by~\cref{equ_ha}, Attention also denotes SW-MSA/W-MSA, $\alpha$ is a hyperparameter.

    \begin{equation}\label{equ_ha}
        \begin{aligned}
       \operatorname{HA}(X) = \operatorname{Attention}(X) + \alpha \operatorname{CAB}(X).\\
        \end{aligned}
    \end{equation}

\section{Experimental Results and Analysis}
\label{sec:exp}
\subsection{Datasets}
\begin{table*}[!htb]
\centering
\caption{Quantitative results achieved by different methods on the KITTI 2012 \cite{geiger2012we}, KITTI 2015 \cite{geiger2015kitti}, Middlebury \cite{scharstein2014high}, and Flickr1024 \cite{wang2019flickr1024} datasets. The best results are in \textbf{bold faces} and the second results are in \underline{underline}. The results of \textcolor[rgb]{0.47,0.47,0.47}{gray color} indicates that the data ensemble strategy was adopted.
} \label{tab:flickr1024_kitti_mid}

\resizebox{\textwidth}{!}
{
\begin{tabular}{lccccccccc}
\toprule
\multirow{2}*{Methods} & \multirow{2}*{Scale} & \multirow{2}*{Tunable Param} & \multicolumn{3}{c}{\textit{Left} }& \multicolumn{4}{c}{$\left(\textit{Left}+\textit{Right} \right)/2$ }\\
\cmidrule(lr){4-6} \cmidrule(lr){7-10}
         &      &           & KITTI 2012 & KITTI 2015 & Middlebury & KITTI 2012 & KITTI 2015 & Middlebury & Flickr1024\\
\midrule
VDSR~\cite{kim2016accurate} & $\times$2 & 0.66M & 30.17$/$0.9062 & 28.99$/$0.9038 & 32.66$/$0.9101 & 30.30$/$0.9089 & 29.78$/$0.9150& 32.77$/$0.9102 & 25.60$/$0.8534\\
EDSR~\cite{lim2017enhanced} & $\times$2 & 38.6M & 30.83$/$0.9199 & 29.94$/$0.9231 & 34.84$/${0.9489} &30.96$/$0.9228 & 30.73$/$0.9335 & {34.95}$/${0.9492} & {28.66}$/$0.9087 \\
RDN~\cite{zhang2018residual} & $\times$2 & 22.0M  & 30.81$/$0.9197 & 29.91$/$0.9224 & {34.85}$/$0.9488 &30.94$/$0.9227 & 30.70$/$0.9330 & 34.94$/$0.9491 & 28.64$/$0.9084 \\
RCAN~\cite{zhang2018image} & $\times$2 & 15.3M & 30.88$/$0.9202 & 29.97$/$0.9231 & 34.80$/$0.9482 & 31.02$/$0.9232 & 30.77$/$0.9336 & 34.90$/$0.9486 & 28.63$/$0.9082 \\
SwinIR~\cite{liang2021swinir} & $\times$2 & 11.75M & 31.13$/$0.9248 & 30.84$/$0.9347 & 35.50$/$0.9535 & 31.24$/$0.9263 & 30.94$/$0.9361 & 35.62$/$0.9533 & 28.84$/$0.9120 \\

HAT-L~\cite{chen2023activating} & $\times$2 & 40.70M & {31.42}$/${0.9275} & {31.14}$/${0.9374} & {36.25}$/${0.9571} & {31.53}$/${0.9289} & {31.23}$/${0.9387} & {36.32}$/${0.9569} & {29.54}$/${0.9189} \\
\midrule
StereoSR~\cite{jeon2018enhancing} & $\times$2 &1.08M & 29.42$/$0.9040 & 28.53$/$0.9038 & 33.15$/$0.9343 & 29.51$/$0.9073 & 29.33$/$0.9168 & 33.23$/$0.9348 & 25.96$/$0.8599 \\
PASSRnet~\cite{wang2019learning} & $\times$2 & 1.37M & 30.68$/$0.9159 & 29.81$/$0.9191 & 34.13$/$0.9421 & 30.81$/$0.9190 & 30.60$/$0.9300 & 34.23$/$0.9422 & 28.38$/$0.9038 \\
iPASSR~\cite{wang2021symmetric} & $\times$2 & 1.37M & {30.97}$/${0.9210} & {30.01}$/${0.9234} & 34.41$/$0.9454 & {31.11}$/${0.9240} & {30.81}$/${0.9340} & 34.51$/$0.9454 & 28.60$/${0.9097} \\
SSRDE-FNet~\cite{dai2021feedback}   & $\times$2 & 2.10M & {31.08}$/${0.9224} & {30.10}$/${0.9245} & {35.02}$/${0.9508} & {31.23}$/${0.9254} & {30.90}$/${0.9352} & {35.09}$/${0.9511} & {28.85}$/${0.9132} \\
NAFSSR-B~\cite{chu2022nafssr} & $\times$2 & 6.77M  & 31.40$/$0.9254 & 30.42$/${0.9282} & {35.62}$/${0.9545} & {31.55}$/${0.9283} & {31.22}$/${0.9380} & {35.68}$/${0.9544} & {29.54}$/${0.9204} \\
NAFSSR-L~\cite{chu2022nafssr} & $\times$2 & 23.79M  & {31.45}$/$0.9261 & 30.46$/$0.9289 & 35.83$/$0.9559 & {31.60}$/${0.9291} & {31.25}$/$0.9386 & 35.88$/$0.9557 & {29.68}$/${0.9221} \\
Steformer~\cite{lin2023steformer} & $\times$2 & 1.29M  & 31.16$/$0.9236 & 30.27$/$0.9271 & 35.15$/$0.9512 & 31.29$/$0.9263 & 31.07$/$0.9371 & 35.23$/$0.9511 & 28.97$/$0.9141 \\
ASteISR (\textbf{Ours}) & $\times$2 & 1.97M  & \underline{31.68}$/$\underline{0.9304} & \underline{31.32}$/$\underline{0.9393} & \underline{36.39}$/$\underline{0.9598} & \underline{31.79}$/$\underline{0.9317} & \underline{31.41}$/$\underline{0.9405} & \underline{36.45}$/$\underline{0.9594} & \underline{30.22}$/$\underline{0.9281} \\
\rowcolor[HTML]{F0F0F0}
ASteISR (\textbf{Ours}) & $\times$2 & 1.97M  & \textbf{31.76}$/$\textbf{0.9310} & \textbf{31.38}$/$\textbf{0.9399} & \textbf{36.54}$/$\textbf{0.9605} & \textbf{31.86}$/$\textbf{0.9323} & \textbf{31.48}$/$\textbf{0.9411} & \textbf{36.60}$/$\textbf{0.9601} & \textbf{30.33}$/$\textbf{0.9290} \\

\midrule
\midrule
VDSR~\cite{kim2016accurate} &  $\times$4 & 0.66M & 25.54$/$0.7662 & 24.68$/$0.7456 & 27.60$/$0.7933 & 25.60$/$0.7722 & 25.32$/$0.7703 & 27.69$/$0.7941 & 22.46$/$0.6718 \\
EDSR~\cite{lim2017enhanced} &  $\times$4 & 38.9M & 26.26$/$0.7954 & 25.38$/$0.7811 & 29.15$/${0.8383} & 26.35$/$0.8015 & 26.04$/$0.8039 & 29.23$/$0.8397 & 23.46$/$0.7285 \\
RDN~\cite{zhang2018residual} &  $\times$4 & 22.0M  & 26.23$/$0.7952 & 25.37$/$0.7813 & 29.15$/$0.8387 & 26.32$/$0.8014 & 26.04$/$0.8043 & 29.27$/${0.8404} & 23.47$/${0.7295} \\
RCAN~\cite{zhang2018image} &  $\times$4 & 15.4M & 26.36$/$0.7968 & 25.53$/$0.7836 & {29.20}$/$0.8381 & 26.44$/$0.8029 & 26.22$/$0.8068 & {29.30}$/$0.8397 & {23.48}$/$0.7286 \\
SwinIR~\cite{liang2021swinir} & $\times$4 & 11.90M & 26.65$/$0.8084 & 26.46$/$0.8135 & 29.55$/$0.8456 & 26.75$/$0.8115 & 26.53$/$0.8161 & 29.69$/$0.8463 & 23.74$/$0.7407 \\

HAT-L~\cite{chen2023activating} & $\times$4 & 40.84M & 26.90$/$0.8148 & {26.75}$/${0.8212} & \underline{30.49}$/${0.8675} & 27.00$/$0.8177 & 26.83$/$0.8238 & \underline{30.65}$/${0.8672} & 24.21$/$0.7590 \\
\midrule
StereoSR~\cite{jeon2018enhancing}  &  $\times$4 & 1.42M   & 24.49$/$0.7502 & 23.67$/$0.7273 &27.70$/$0.8036 & 24.53$/$0.7555 & 24.21$/$0.7511 & 27.64$/$0.8022 & 21.70$/$0.6460 \\
PASSRnet~\cite{wang2019learning}  &  $\times$4 & 1.42M   & 26.26$/$0.7919 & 25.41$/$0.7772 &28.61$/$0.8232 & 26.34$/$0.7981 & 26.08$/$0.8002 & 28.72$/$0.8236 & 23.31$/$0.7195 \\
SRRes+SAM~\cite{ying2020stereo}  &  $\times$4 & 1.73M  & 26.35$/$0.7957 & 25.55$/$0.7825 & 28.76$/$0.8287 & 26.44$/$0.8018 & 26.22$/$0.8054 & 28.83$/$0.8290 & 23.27$/$0.7233 \\
iPASSR~\cite{wang2021symmetric}  &  $\times$4 & 1.42M  & {26.47}$/${0.7993} & {25.61}$/${0.7850} & 29.07$/$0.8363 & {26.56}$/${0.8053} & {26.32}$/${0.8084} & 29.16$/$0.8367 & 23.44$/$0.7287 \\
SSRDE-FNet~\cite{dai2021feedback}  & $\times$4 & 2.24M  & {26.61}$/${0.8028} & {25.74}$/${0.7884} & {29.29}$/${0.8407} & {26.70}$/${0.8082} & {26.43}$/${0.8118} & {29.38}$/${0.8411} & {23.59}$/${0.7352} \\
NAFSSR-B \cite{chu2022nafssr} & $\times$4 & 6.80M  & {26.99}$/${0.8121} & {26.17}$/${0.8020} & {29.94}$/${0.8561} & {27.08}$/${0.8181} & {26.91}$/${0.8245} & {30.04}$/${0.8568} & {24.07}$/${0.7551} \\

NAFSSR-L \cite{chu2022nafssr} & $\times$4 & 23.83M  & {27.04}$/${0.8135} & {26.22}$/${0.8034} & {30.11}$/${0.8601} & {27.12}$/${0.8194} & {26.96}$/${0.8257} & {30.18}$/${0.8596} & {24.17}$/${0.7589} \\
Steformer~\cite{lin2023steformer} & $\times$4 & 1.34M  & 26.61$/$0.8037 & 25.74$/$0.7906 & 29.29$/$0.8424 & 26.70$/$0.8098 & 26.45$/$0.8134 & 29.38$/$0.8425 & 23.58$/$0.7376 \\
SCGLANet~\cite{zhou2023stereo} & $\times$4 & 25.29M  &27.03$/$0.8154 & 26.18$/$0.8052 &30.23$/$0.8627 & {27.10}$/${0.8209} & {26.87}$/${0.8263} & {30.04}$/${0.8568} & {24.30}$/${0.7657} \\
\rowcolor[HTML]{F0F0F0}

HTCAN~\cite{cheng2023hybrid} & $\times$4 & 64.82M  & \textbf{27.16}$/${0.8189} & {26.26}$/${0.8083} & {30.25}$/${0.8628} & \underline{27.25}$/$\textbf{0.8249} & \underline{26.99}$/$\underline{0.8299} & {30.33}$/${0.8634} & \underline{24.44}$/$\underline{0.7703} \\

ASteISR (\textbf{Ours}) & $\times$4 & 1.97M  & {27.07}$/$\underline{0.8203} & \underline{26.91}$/$\underline{0.8266} & {30.48}$/$\underline{0.8687} & {27.17}$/${0.8231} & {26.98}$/$\underline{0.8299} & {30.63}$/$\underline{0.8701} & {24.43}$/${0.7690} \\
\rowcolor[HTML]{F0F0F0}
ASteISR (\textbf{Ours}) & $\times$4 & 1.97M  & \underline{27.15}$/$\textbf{0.8218} & \textbf{27.00}$/$\textbf{0.8283} & \textbf{30.61}$/$\textbf{0.8707} & \textbf{27.26}$/$\underline{0.8245} & \textbf{27.07}$/$\textbf{0.8305} & \textbf{30.75}$/$\textbf{0.8720} & \textbf{24.52}$/$\textbf{0.7710} \\

\bottomrule

\end{tabular}}
\vspace{-3mm}
\end{table*}

\textbf{Training Dataset:} The training datasets used in this study were Flickr1024~\cite{wang2019flickr1024} and Middlebury~\cite{scharstein2014high}, which are commonly employed in SteISR research~\cite{chu2022nafssr}. The combined training dataset consists of 860 pairs of stereo images, with Flickr1024 contributing 800 pairs and Middlebury providing 60 pairs.

\textbf{Testing Dataset:} The model's performance was evaluated using four widely used datasets during the testing phase: KITTI2012~\cite{geiger2012we}, KITTI2015~\cite{geiger2015kitti}, Middlebury~\cite{scharstein2014high}, and Flickr1024~\cite{wang2019flickr1024}. These datasets include 20, 20, 5, and 112 pairs of stereo images, respectively.

\subsection{Implementation Details}

We cropped the input images into small patches sized 128x320 for training purposes. Initially, the pre-training weights of HAT were loaded, specifically utilizing the HAT-L model structure pre-trained on ImageNet. During training, the parameters of HAT-L remained frozen, while only the parameters of the stereo adapter and spatial adapter were updated. The training loss function was set as L1, and the optimizer was set as AdamW. Notably, the initial learning rate was set to a relatively large value of 5e-4. The model was fine-tuned on two A40 GPUs with a batch size of 3. A total of 200,000 iterations were conducted for fine-tuning. Finally, the fine-tuned model was evaluated using PSNR and SSIM metrics.

\subsection{Comparison with State-of-the-Art Methods}

\textbf{Quantitative results.} Initially, a detailed comparison was made with the state-of-the-art SISR methods, including  VDSR~\cite{kim2016accurate}, EDSR \cite{lim2017enhanced}, RDN~\cite{zhang2018residual}, RCAN~\cite{zhang2018image}, SwinIR~\cite{liang2021swinir}, and HAT~\cite{chen2023activating}. Subsequently, the comparison shifts to the state-of-the-art SteISR methods, which include StereoSR~\cite{jeon2018enhancing}, PASSRNet~\cite{wang2019learning}, SRRes+SAM~\cite{ying2020stereo}, iPASSR~\cite{wang2021symmetric}, SSRDE-FNet~\cite{dai2021feedback}, SwinIPASSR~\cite{jin2022swinipassr}, NAFSSR~\cite{chu2022nafssr}, SCGLANet~\cite{zhou2023stereo}, Steformer~\cite{lin2023steformer} and HTCAN~\cite{cheng2023hybrid}. The comparison results are summarized in \cref{tab:flickr1024_kitti_mid}. \textbf{It is worth mentioning that the papers on HTCAN~\cite{cheng2023hybrid} and SCGLANet~\cite{zhou2023stereo} only include results for x4, so the results for x2 were not added to~\cref{tab:flickr1024_kitti_mid}.}
\begin{figure}[htb]
\scriptsize
\centering
\begin{tabular}{cc}
\hspace{-0.4cm}
\begin{adjustbox}{valign=t}
\begin{tabular}{c}
\includegraphics[width=0.158\textwidth,height=0.135\textheight]{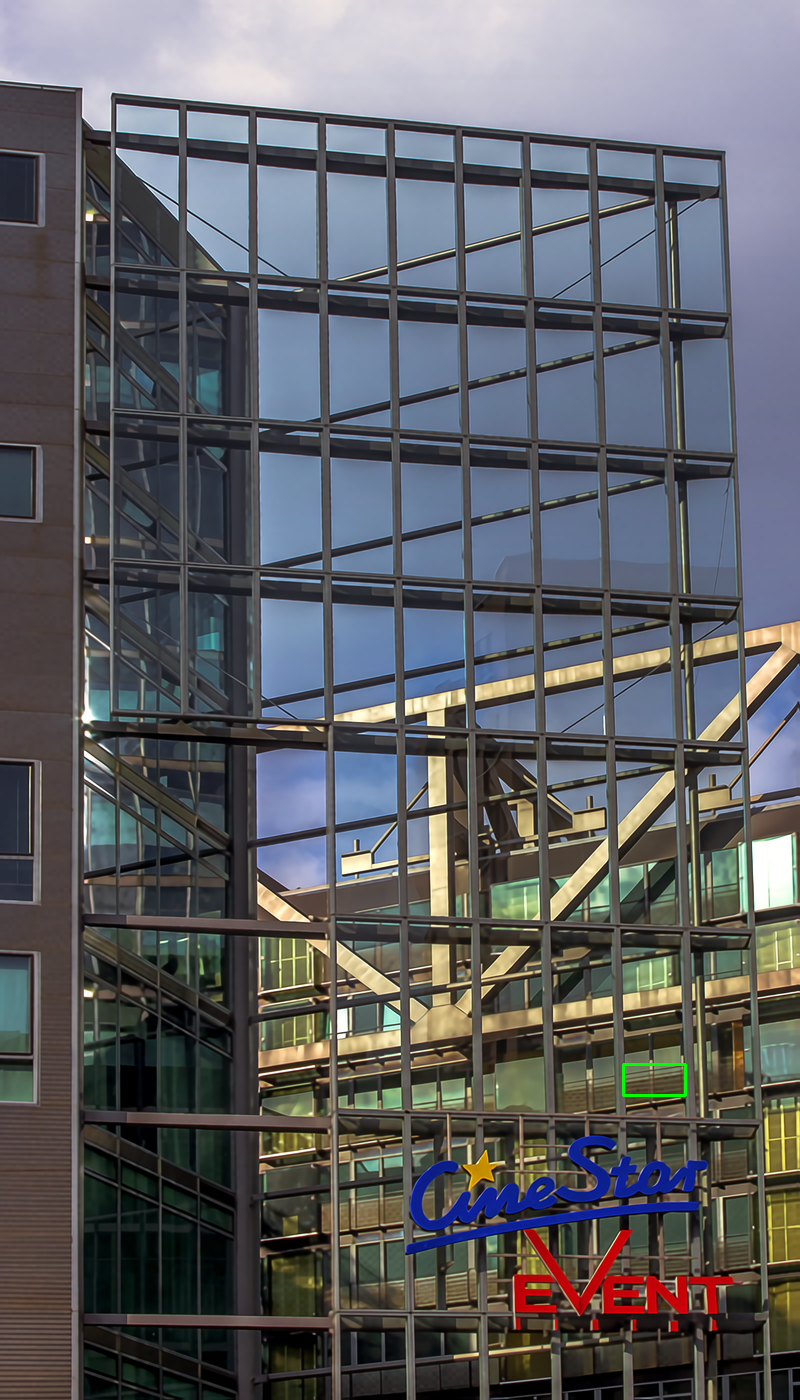}
\\
img\_0010
\end{tabular}
\end{adjustbox}
\hspace{-0.26cm}
\begin{adjustbox}{valign=t}
\begin{tabular}{cccccc}
\includegraphics[width=0.180\textwidth]{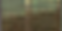} \hspace{-4mm} &
\includegraphics[width=0.180\textwidth]{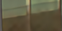} \hspace{-4mm} &
\includegraphics[width=0.180\textwidth]{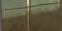} \hspace{-4mm} &
\includegraphics[width=0.180\textwidth]{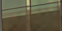} \hspace{-4mm} &
\includegraphics[width=0.180\textwidth]{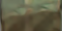} \hspace{-4mm}
\\

\tiny Bicubic \hspace{-4mm} &
\tiny RCAN~\cite{zhang2018image} \hspace{-4mm} &
\tiny SwinIR~\cite{liang2021swinir} \hspace{-4mm} &
\tiny HAT-L~\cite{chen2023activating} \hspace{-4mm} &
\tiny StereoSR~\cite{jeon2018enhancing} \hspace{-4mm} &
\\

\includegraphics[width=0.180\textwidth]{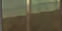} \hspace{-4mm} &
\includegraphics[width=0.180\textwidth]{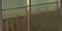} \hspace{-4mm} &
\includegraphics[width=0.180\textwidth]{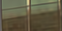} \hspace{-4mm} &
\includegraphics[width=0.180\textwidth]{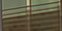} \hspace{-4mm}   &
\includegraphics[width=0.180\textwidth]{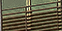} \hspace{-4mm}
\\

\tiny iPASSR~\cite{wang2021symmetric} \hspace{-4mm} &
\tiny NAFSSR-L~\cite{chu2022nafssr}  \hspace{-4mm} &
\tiny SCGLANet~\cite{zhou2023stereo}  \hspace{-4mm} &
\tiny ASteISR (\textbf{Ours})  \hspace{-4mm} &
Ground Truth \hspace{-4mm}
\\
\end{tabular}
\end{adjustbox}
\vspace{1mm}
\\

\hspace{-0.4cm}
\begin{adjustbox}{valign=t}
\begin{tabular}{c}
\includegraphics[width=0.158\textwidth,height=0.170\textheight]{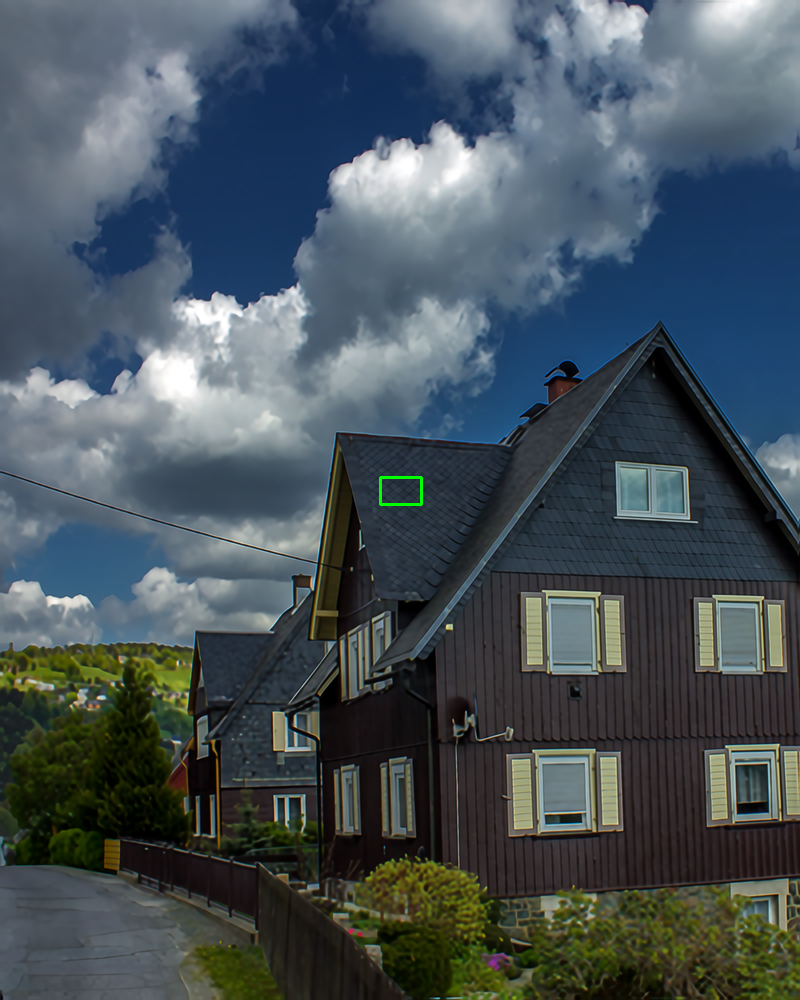}
\\
img\_0041
\end{tabular}
\end{adjustbox}
\hspace{-0.26cm}
\begin{adjustbox}{valign=t}
\begin{tabular}{cccccc}

\includegraphics[width=0.180\textwidth]{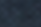} \hspace{-4mm} &
\includegraphics[width=0.180\textwidth]{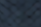} \hspace{-4mm} &
\includegraphics[width=0.180\textwidth]{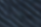} \hspace{-4mm} &
\includegraphics[width=0.180\textwidth]{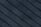} \hspace{-4mm} &
\includegraphics[width=0.180\textwidth]{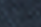} \hspace{-4mm}
\\

\tiny Bicubic \hspace{-4mm} &
\tiny RCAN~\cite{zhang2018image} \hspace{-4mm} &
\tiny SwinIR~\cite{liang2021swinir} \hspace{-4mm} &
\tiny HAT-L~\cite{chen2023activating} \hspace{-4mm} &
\tiny StereoSR~\cite{jeon2018enhancing} \hspace{-4mm} &
\\

\includegraphics[width=0.180\textwidth]{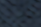} \hspace{-4mm} &
\includegraphics[width=0.180\textwidth]{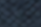} \hspace{-4mm} &
\includegraphics[width=0.180\textwidth]{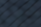} \hspace{-4mm} &
\includegraphics[width=0.180\textwidth]{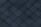} \hspace{-4mm}   &
\includegraphics[width=0.180\textwidth]{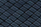} \hspace{-4mm}
\\

\tiny iPASSR~\cite{wang2021symmetric} \hspace{-4mm} &
\tiny NAFSSR-L~\cite{chu2022nafssr}  \hspace{-4mm} &
\tiny SCGLANet~\cite{zhou2023stereo}  \hspace{-4mm} &
\tiny ASteISR (\textbf{Ours})  \hspace{-4mm} &
\tiny Ground Truth \hspace{-4mm}
\\
\end{tabular}
\end{adjustbox}
\vspace{1mm}
\\
\end{tabular}
\caption{Visual results ($\times$4) achieved by different methods on the  Flickr1024~\cite{wang2019learning} dataset.
}
\label{fig:flickr1024}
\end{figure} 
\begin{figure*}[!t]
\hspace{-0.3cm}
\scalebox{0.98}{
\begin{tabular}[b]{ccccccc }

    \includegraphics[width=.138\textwidth,valign=t]{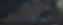} &
    \includegraphics[width=.138\textwidth,valign=t]{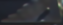} &
    \includegraphics[width=.138\textwidth,valign=t]{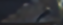} &
    \includegraphics[width=.138\textwidth,valign=t]{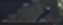} &
    \includegraphics[width=.138\textwidth,valign=t]{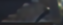} &
    \includegraphics[width=.138\textwidth,valign=t]{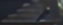} &
    \includegraphics[width=.138\textwidth,valign=t]{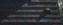}
   \\
   \vspace{0.1mm}
   \\

 \includegraphics[width=.138\textwidth,valign=t]{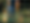} &
 \includegraphics[width=.138\textwidth,valign=t]{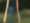} &
 \includegraphics[width=.138\textwidth,valign=t]{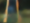} &
 \includegraphics[width=.138\textwidth,valign=t]{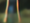} &
 \includegraphics[width=.138\textwidth,valign=t]{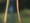} &
 \includegraphics[width=.138\textwidth,valign=t]{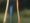} &
 \includegraphics[width=.138\textwidth,valign=t]{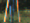}
  \\

 \vspace{0.1mm}
 \\

 \includegraphics[width=.138\textwidth,valign=t]{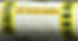} &
 \includegraphics[width=.138\textwidth,valign=t]{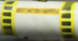} &
 \includegraphics[width=.138\textwidth,valign=t]{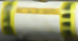} &
 \includegraphics[width=.138\textwidth,valign=t]{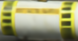} &
 \includegraphics[width=.138\textwidth,valign=t]{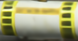} &
 \includegraphics[width=.138\textwidth,valign=t]{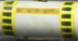} &
 \includegraphics[width=.138\textwidth,valign=t]{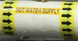} \\

\tiny~Bicubic    &
\tiny~HAT-L~\cite{chen2023activating} &
\tiny~iPASSR~\cite{wang2021symmetric} & \tiny~NAFSSR-L ~\cite{chu2022nafssr} &\tiny~SCGLANet~\cite{zhou2023stereo} &\tiny~ASteISR(\textbf{Ours})  & \tiny~Ground Truth\\

\end{tabular}
}
\vspace{-1mm}
\caption{Visual results ($\times$4) achieved by different methods on the KITTI 2012~\cite{geiger2012we} (top), KITTI 2015~  \cite{geiger2015kitti} (center) and Middlebury \cite{scharstein2014high} (bottom) dataset.
}
\label{fig:kittimb}
\vspace{-3mm}
\end{figure*}

According to the results presented in~\cref{tab:flickr1024_kitti_mid}, our method, which fine-tuned a mere 1.97M model parameters, achieved outstanding performance on nearly all test data. Specifically, for 2x super-resolution, our method surpassed the HAT-L~\cite{chen2023activating} single-image super-resolution model, which trained 40.7M parameters, by only fine-tuning approximately \textbf{$4.8\%$} of its total parameters. On the Flickr1024 dataset, our model attained a PSNR of 30.33dB, surpassing it by an impressive \textbf{0.8dB}. Similarly, when compared to the NAFSSR-L~\cite{chu2022nafssr} stereo image super-resolution model with 23.79M parameters, we required training of approximately \textbf{$8.2\%$} of its model parameters. On the Flickr1024 dataset, our model even outperformed it by \textbf{0.65dB}.

For 4x super-resolution, compared to HAT-L~\cite{chen2023activating}, our approach surpassed it by 0.31dB on the Flickr1024 dataset. In comparison to the larger two-stage stereo super-resolution HTCAN~\cite{cheng2023hybrid} model, even though we fine-tuned only $3.0\%$ of its parameters, our model achieved competitive results on all datasets. Particularly on the Middlebury dataset, our model outperformed HTCAN by \textbf{0.42dB}. This strongly demonstrates the effectiveness and efficiency of our proposed method.

\textbf{Qualitative results.} Further qualitative evaluation results of our approach compared to other methods are shown in \cref{fig:kittimb} and \cref{fig:flickr1024}. The figures clearly demonstrate that our method can accurately reconstruct textures that are more natural, realistic, and complete compared to other single-image and stereo super-resolution methods. This substantiates the effectiveness of our method.

\subsection{Ablation Study}
\label{sec:ab}

\begin{figure}[htbp]
    \centering
    \begin{minipage}[b]{0.56\textwidth}
        \centering
       \includegraphics[width=\textwidth,height=0.6\textwidth]{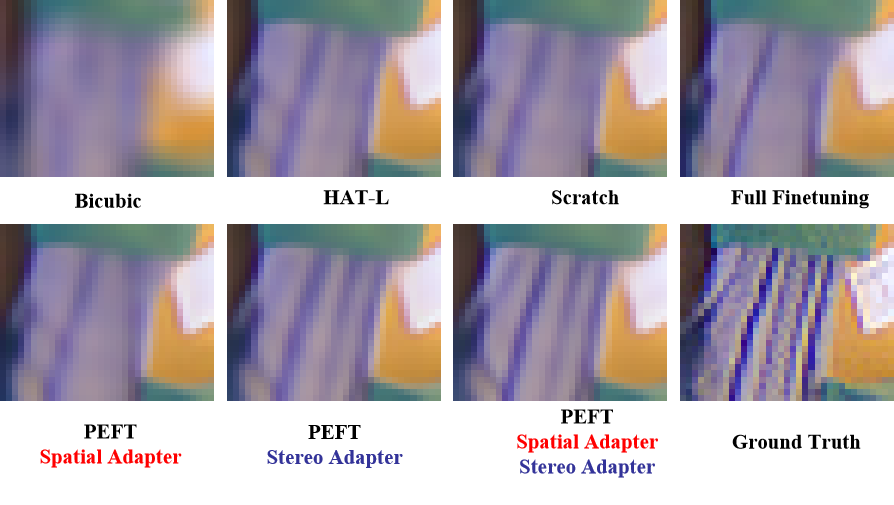}
    \caption{Visual results of different fine-tuning methods.}
        \label{fig:difpeft}
    \end{minipage}
     \hfill
    \begin{minipage}[b]{0.42\textwidth}
        \centering
       \includegraphics[width=0.98\textwidth]{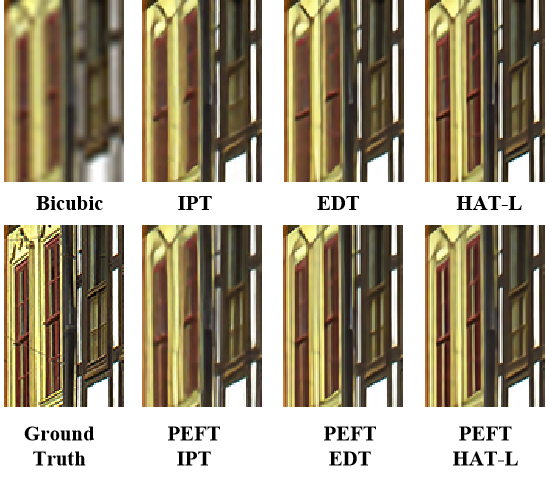}
        \caption{Visual results of finetuning on different pre-training models.}
          \label{fig:diffpretrained}
    \end{minipage}

\end{figure}

\textbf{Traditional Finetuning VS PEFT.} We utilized HAT-L~\cite{chen2023activating} as the base model and studied the impact of different fine-tuning methods on the performance of SteISR. These methods included training from scratch, full fine-tuning, and the PEFT method. Quantitative results can be found in~\cref{tab:abl_adapter}, where HAT-L (Frozen) denotes the results achieved by directly using the original HAT-L model for inference. From~\cref{tab:abl_adapter}, it is apparent that both training from scratch and full fine-tuning methods yield inferior performance compared to the original HAT-L model. In terms of PEFT, when fine-tuning the spatial adapter, the model's performance, although not surpassing that of the original HAT-L model, only requires $42\%$ of the fine-tuning time and $85\%$ of the storage consumption when compared to full fine-tuning. If the stereo adapter is fine-tuned, the fine-tuning time and memory consumption are further reduced, while the model's performance surpasses that of both full fine-tuning and the original HAT-L model. Moreover, incorporating the spatial adapter on top of fine-tuning the stereo adapter leads to further improvements in the model's performance. \cref{fig:difpeft} further illustrates the qualitative results of different fine-tuning methods. As shown in \cref{fig:difpeft}, it can prove that the model trained on large dataset (HAT-L) has better priors to restore texture compared to the model trained on small dataset (Scratch). In addition, the figure also shows the ``catastrophic forgetting'' using small dataset to fully finetune teh large pre-trained model, which deficits in texture recovering ability after full-finetuning. Another clear evidence suggests that adding a stereo adapter can effectively improve the performance of the model. Finally, combining stereo adapters with spatial adapters can further promote the performance of model for texture repairing.

\begin{table}[!htb]
\centering
\caption{Results of different fine-tuning methods. The best results are in \textbf{bold faces} and the second results are in \underline{underline}. The $\textcolor{darkpink}{ \blacktriangledown}$ denotes deterioration, and the $\textcolor{darkpastelgreen}{ \blacktriangle}$ denotes imporvement compared to Frozen. } 
\label{tab:abl_adapter}
\resizebox{0.85\textwidth}{!}
{
\begin{tabular}{ccccc}
\hline
\multirow{2}{*}{Methods} & \multirow{2}{*}{\begin{tabular}[c]{@{}c@{}}Tunable Params\\ (M)\end{tabular}} & \multirow{2}{*}{\begin{tabular}[c]{@{}c@{}}Time \\ (H)\end{tabular}} & \multirow{2}{*}{\begin{tabular}[c]{@{}c@{}}Mem \\ (G)\end{tabular}} & \multirow{2}{*}{PSNR/SSIM} \\
 &  &  &  &  \\ \hline
HAT-L(Frozen) & 0 & 0 & 0 & 24.21/0.7590 \\ \hline
Scratch & 40.85 & 127 & 25.11 & 23.78$\textcolor{darkpink}{\scriptstyle \blacktriangledown0.43}$/0.7410$\textcolor{darkpink}{\scriptstyle \blacktriangledown0.0181}$ \\
Full Finetuning & 40.85 & 127 & 25.11 & 24.18$\textcolor{darkpink}{\scriptstyle \blacktriangledown0.03}$/0.7517$\textcolor{darkpink}{\scriptstyle \blacktriangledown0.0073}$ \\
PEFT \color[HTML]{FE0000}{Spatial} Adapter & \underline{1.18} & \underline{54} & \underline{21.32} & 24.20$\textcolor{darkpink}{\scriptstyle \blacktriangledown0.01}$/0.7523$\textcolor{darkpink}{\scriptstyle \blacktriangledown0.0067}$ \\
PEFT \color[HTML]{0000FE}{Stereo} Adapter & \textbf{0.78} & \textbf{50} & \textbf{17.54} & \underline{24.33$\textcolor{darkpastelgreen}{\scriptstyle \blacktriangle0.12}$/0.7636$\textcolor{darkpastelgreen}{\scriptstyle \blacktriangle0.0046}$} \\
PEFT \color[HTML]{FE0000}{Spatial}+\color[HTML]{0000FE}{Stereo} Adapter & 1.97 & 55 & 21.37 & \textbf{24.43$\textcolor{darkpastelgreen}{\scriptstyle \blacktriangle0.22}$/0.7690$\textcolor{darkpastelgreen}{\scriptstyle \blacktriangle0.0100}$} \\ \hline
\end{tabular}
}
\end{table}

\textbf{Effects of Pre-training Datasets.} The impact of pre-trained models using different datasets on the PEFT method is presented in Table \ref{tab:abl_diffdata}. The results clearly demonstrate that models pre-trained ones on larger datasets, such as ImageNet, outperform those pre-trained on smaller datasets like DF2K.

\textbf{Different Pre-training Models.} To examine the performance of the proposed PEFT method with various pre-trained models, We perform PEFT on representative pre-trained models on large datasets, including EDT~\cite{li2021efficient} and IPT~\cite{chen2021pre}. The quantitative results, presented in~\cref{tab:abl_difpre}, clearly demonstrate significant improvements achieved by the models that were finetuned using the PEFT method. The qualitative comparison before and after using PEFT was shown in \cref{fig:diffpretrained}, which provides clear evidence that all models that have undergone PEFT have achieved an improvement in texture repair ability. This confirms the versatility and effectiveness of our proposed method across different pre-training models.

\begin{table}[htbp]
    \centering
    \begin{minipage}[b]{0.5\linewidth}
        \centering
        \caption{Results on different pre-training dataset. The $\textcolor{darkpastelgreen}{ \blacktriangle}$ denotes imporvement compared to the original one.} 
        \label{tab:abl_diffdata}
        \resizebox{\textwidth}{!}{
            \begin{tabular}{cccc}
                \hline
                \multicolumn{1}{l}{\begin{tabular}[c]{@{}c@{}}Pre-trained\\ Model\end{tabular}} & \multicolumn{1}{c}{\begin{tabular}[c]{@{}c@{}}Pre-trained\\ Dataset\end{tabular}} & \multicolumn{1}{c}{\begin{tabular}[c]{@{}c@{}}Orig\\ (PSNR/SSIM)\end{tabular}} & \multicolumn{1}{c}{\begin{tabular}[c]{@{}c@{}}Tuned\\ (PSNR/SSIM)\end{tabular}} \\ \hline
                \multirow{4}{*}{EDT} & DF2K & 23.87/0.7455 & 24.00$\textcolor{darkpastelgreen}{\scriptstyle \blacktriangle0.13}$/0.7534$\textcolor{darkpastelgreen}{\scriptstyle \blacktriangle0.0079}$ \\
                 & ImageNet & 23.86/0.7444 & 24.01$\textcolor{darkpastelgreen}{\scriptstyle \blacktriangle0.15}$/0.7534$\textcolor{darkpastelgreen}{\scriptstyle \blacktriangle0.0090}$ \\
                 & \multicolumn{1}{c}{\begin{tabular}[c]{@{}c@{}}ImageNet\\ +DF2K\end{tabular}} & 23.93/0.7469 & 24.13$\textcolor{darkpastelgreen}{\scriptstyle \blacktriangle0.20}$/0.7562$\textcolor{darkpastelgreen}{\scriptstyle \blacktriangle0.0093}$ \\ \hline
            \end{tabular}
            }
        
    \end{minipage}
    \hfill 
    \begin{minipage}[b]{0.48\linewidth}
      
         \centering
        \caption{Results on different pre-training model. The $\textcolor{darkpastelgreen}{ \blacktriangle}$ denotes imporvement compared to the original one.} 
        \label{tab:abl_difpre}
        \resizebox{\textwidth}{!}{
            \begin{tabular}{ccc}
                \hline
                \multicolumn{1}{l}{\begin{tabular}[c]{@{}c@{}}Pre-trained\\ Model\end{tabular}} & \multicolumn{1}{c}{\begin{tabular}[c]{@{}c@{}}Orig\\ (PSNR/SSIM)\end{tabular}} & \multicolumn{1}{c}{\begin{tabular}[c]{@{}c@{}}Tuned\\ (PSNR/SSIM)\end{tabular}} \\ \hline
                IPT & 23.68/0.7366 & 23.87$\textcolor{darkpastelgreen}{\scriptstyle \blacktriangle0.19}$/0.7451$\textcolor{darkpastelgreen}{\scriptstyle \blacktriangle0.0085}$ \\
                EDT & 23.93/0.7469 & 24.13$\textcolor{darkpastelgreen}{\scriptstyle \blacktriangle0.20}$/0.7562$\textcolor{darkpastelgreen}{\scriptstyle \blacktriangle0.0093}$  \\
                HAT\-L & 24.21/0/7590 & 24.43$\textcolor{darkpastelgreen}{\scriptstyle \blacktriangle0.22}$/0.7690$\textcolor{darkpastelgreen}{\scriptstyle \blacktriangle0.0100}$ \\ \hline
            \end{tabular}
        }
    \end{minipage}
\end{table}

\begin{figure}[h]
   \vspace{-0.3cm}
  \begin{minipage}[t]{0.58\textwidth}
        \vspace{-2.4cm}
        \captionsetup{type=table}
        \caption{Results on different position of adapter. The $\textcolor{darkpink}{ \blacktriangledown}$ denotes deterioration.} \label{tab:abl_diffpos}
        \resizebox{1.0\textwidth}{!}
        {
        \begin{tabular}{ccccc}
        \hline
        \multirow{2}{*}{Method} & \multicolumn{1}{c}{\multirow{2}{*}{\begin{tabular}[c]{@{}c@{}}Tunable Params\\ (M)\end{tabular}}} & \multicolumn{1}{c}{\multirow{2}{*}{\begin{tabular}[c]{@{}c@{}}Time \\ (H)\end{tabular}}} & \multicolumn{1}{c}{\multirow{2}{*}{\begin{tabular}[c]{@{}c@{}}Mem \\ (G)\end{tabular}}} & \multirow{2}{*}{PSNR/SSIM} \\
         & \multicolumn{1}{c}{} & \multicolumn{1}{c}{} & \multicolumn{1}{c}{} &  \\ \hline
        baseline & 1.97 & 55 & 21.37 & 24.43/0.7690 \\ \hline
        (a) & 3.68 & 126 & 24.53 & 24.35$\textcolor{darkpink}{\scriptstyle \blacktriangledown0.08}$/0.7653$\textcolor{darkpink}{\scriptstyle \blacktriangledown0.0037}$ \\
        (b) & 3.68 & 124 & 24.53 & 24.39$\textcolor{darkpink}{\scriptstyle \blacktriangledown0.04}$/0.7661$\textcolor{darkpink}{\scriptstyle \blacktriangledown0.0029}$ \\ \hline
        \end{tabular}
        }
  \end{minipage}
  \hfill
  \begin{minipage}[t]{0.4\textwidth}

     \includegraphics[width=1.05\linewidth]{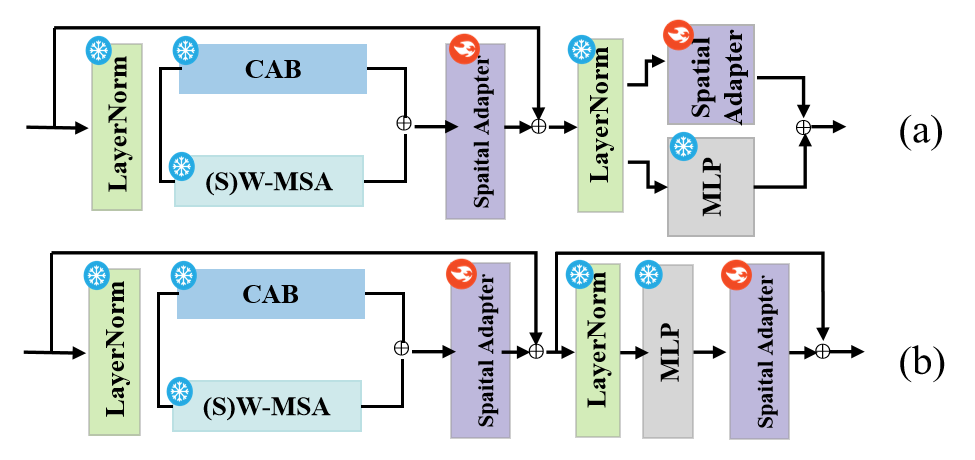}
      \captionsetup{type=figure}
      \vspace{-0.5cm}
    \caption{The difference position of spatial adapters.}
      \label{fig:diffpos}
  \end{minipage}
\end{figure}

\begin{figure}[htbp]
  \begin{minipage}[b]{0.46\linewidth}
    \centering
    \vspace{-0.5cm}
    \includegraphics[width=0.9\linewidth,height=0.7\linewidth]{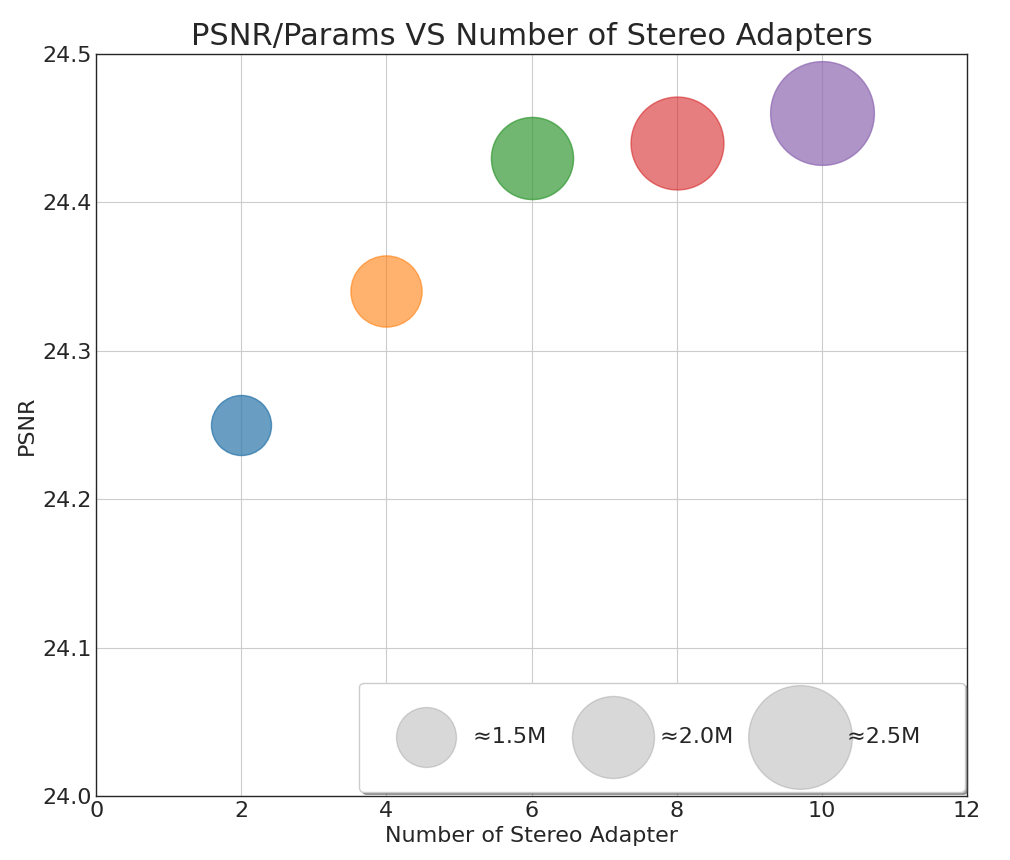}
    \vspace{-0.2cm}
    \caption{The comparison results between model performance and the number of stereo adapters.}
      \label{fig:diffstereo}
      
  \end{minipage}
  \hfill
  \begin{minipage}[t]{0.48\linewidth}
    \centering
    \vspace{-5.0cm}
    \includegraphics[width=0.9\linewidth]{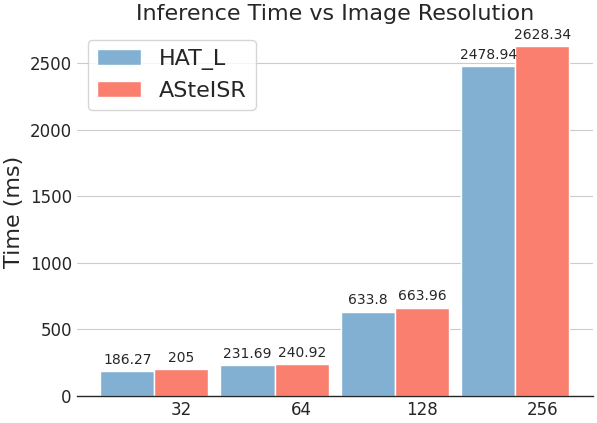}
    \caption{The comparison results of inference time at different resolutions.}
      \label{fig_infertime}
  \end{minipage}
\end{figure}

\textbf{Different Placement of Spatial Adapter and Number of Stereo Adapters.} The influence of spatial adapter placement on the fine-tuning performance of SteISR is shown as \cref{tab:abl_diffpos}, where positions (a) and (b) are referenced to \cref{fig:diffpos}. The table reveals that the additional adapter after the MLP and the parallel placement of adapters both lead to an increase in fine-tuned parameters and fine-tuning time. Consequently, this results in varying degrees of performance degradation. The impact of a different number of stereo adapters is depicted in \cref{fig:diffstereo}. The figure illustrates the comparison of the number of adapters and the fine-tuning results. From the figure, it is evident that as the number of stereo adapters increases, the model performance also improves, eventually reaching saturation.

\textbf{Discussion and Limitations.}  Although there have been previous methods, such as \cite{lei2021deep} and \cite{Ying2020astere}, which have fine-tuned SISR models to adapt to SteISR models, the emergence of larger models poses specific challenges in terms of performance and device resource utilization, as discussed in \cref{sec:ab}. In comparison to these existing methods, our proposed approach is characterized by its simplicity and effectiveness in addressing these challenges, thus offering valuable insights for the advancement of low-level vision tasks. However, the incorporation of new layers into the pre-trained model inevitably leads to a reduction in the inference speed of the model, as shown in  \cref{fig_infertime}.
\section{Conclusion}

In this paper, we proposed a method to enhance the capability of single-image super-resolution (SISR) models for efficiently processing stereo images. The proposed method capitalizes on pre-trained SISR models by integrating specialized stereo and spatial adapters while updating exclusively the parameters of these new components during the training phase. This strategy significantly minimizes training time and memory consumption, yet delivers outstanding results on popular stereo image super-resolution benchmarks. Moreover, the simplicity and flexibility of our approach enable seamless integration with new large pre-trained models, facilitating advancements in both stereo and single-image super-resolution. Furthermore, our method can be extended to various low-level vision tasks such as video deblurring and denoising by adapting pre-trained single-image models to the multi-image domain using carefully designed adapters. For future research, we intend to explore the application of universal adapters in low-level vision and study on parameter-efficient fine-tuning method that do not adversely affect the inference time.

%
%
\bibliographystyle{splncs04}
\bibliography{main}

\begin{thebibliography}{10}
\providecommand{\url}[1]{\texttt{#1}}
\providecommand{\urlprefix}{URL }
\providecommand{\doi}[1]{https://doi.org/#1}

\bibitem{agustsson2017ntire}
Agustsson, E., Timofte, R.: Ntire 2017 challenge on single image super-resolution: Dataset and study. In: Proceedings of the IEEE Conference on Computer Vision and Pattern Recognition Workshops. pp. 126--135 (2017)

\bibitem{ansell2021composable}
Ansell, A., Ponti, E.M., Korhonen, A., Vuli{\'c}, I.: Composable sparse fine-tuning for cross-lingual transfer. arXiv preprint arXiv:2110.07560  (2021)

\bibitem{ba2016layer}
Ba, J.L., Kiros, J.R., Hinton, G.E.: Layer normalization. arXiv preprint arXiv:1607.06450  (2016)

\bibitem{chen2021pre}
Chen, H., Wang, Y., Guo, T., Xu, C., Deng, Y., Liu, Z., Ma, S., Xu, C., Xu, C., Gao, W.: Pre-trained image processing transformer. In: Proceedings of the IEEE/CVF Conference on Computer Vision and Pattern Recognition. pp. 12299--12310 (2021)

\bibitem{chen2020big}
Chen, T., Kornblith, S., Swersky, K., Norouzi, M., Hinton, G.E.: Big self-supervised models are strong semi-supervised learners. Advances in Neural Information Processing Systems  \textbf{33},  22243--22255 (2020)

\bibitem{chenempirical}
Chen, X., Xie, S., He, K.: An empirical study of training self-supervised vision transformers. in 2021 ieee. In: CVF International Conference on Computer Vision. pp. 9620--9629

\bibitem{chen2023activating}
Chen, X., Wang, X., Zhou, J., Qiao, Y., Dong, C.: Activating more pixels in image super-resolution transformer. In: Proceedings of the IEEE/CVF Conference on Computer Vision and Pattern Recognition. pp. 22367--22377 (2023)

\bibitem{cheng2023hybrid}
Cheng, M., Ma, H., Ma, Q., Sun, X., Li, W., Zhang, Z., Sheng, X., Zhao, S., Li, J., Zhang, L.: Hybrid transformer and cnn attention network for stereo image super-resolution. In: Proceedings of the IEEE/CVF Conference on Computer Vision and Pattern Recognition. pp. 1702--1711 (2023)

\bibitem{chu2022nafssr}
Chu, X., Chen, L., Yu, W.: Nafssr: Stereo image super-resolution using nafnet. In: Proceedings of the IEEE/CVF Conference on Computer Vision and Pattern Recognition. pp. 1239--1248 (2022)

\bibitem{chuah2022semantic}
Chuah, W., Tennakoon, R., Hoseinnezhad, R., Suter, D., Bab-Hadiashar, A.: Semantic guided long range stereo depth estimation for safer autonomous vehicle applications. IEEE Transactions on Intelligent Transportation Systems  \textbf{23}(10),  18916--18926 (2022)

\bibitem{cosner2022self}
Cosner, R.K., Rodriguez, I.D.J., Molnar, T.G., Ubellacker, W., Yue, Y., Ames, A.D., Bouman, K.L.: Self-supervised online learning for safety-critical control using stereo vision. In: 2022 International Conference on Robotics and Automation. pp. 11487--11493. IEEE (2022)

\bibitem{dai2021feedback}
Dai, Q., Li, J., Yi, Q., Fang, F., Zhang, G.: Feedback network for mutually boosted stereo image super-resolution and disparity estimation. In: Proceedings of the 29th ACM International Conference on Multimedia. pp. 1985--1993 (2021)

\bibitem{deng2009imagenet}
Deng, J., Dong, W., Socher, R., Li, L.J., Li, K., Fei-Fei, L.: Imagenet: A large-scale hierarchical image database. In: Proceedings of the IEEE/CVF International Conference on Computer Vision. pp. 248--255. IEEE (2009)

\bibitem{dettmers2023qlora}
Dettmers, T., Pagnoni, A., Holtzman, A., Zettlemoyer, L.: Qlora: Efficient finetuning of quantized llms. arXiv preprint arXiv:2305.14314  (2023)

\bibitem{devlin2018bert}
Devlin, J., Chang, M.W., Lee, K., Toutanova, K.: Bert: Pre-training of deep bidirectional transformers for language understanding. arXiv preprint arXiv:1810.04805  (2018)

\bibitem{dosovitskiy2020image}
Dosovitskiy, A., Beyer, L., Kolesnikov, A., Weissenborn, D., Zhai, X., Unterthiner, T., Dehghani, M., Minderer, M., Heigold, G., Gelly, S., et~al.: An image is worth 16x16 words: Transformers for image recognition at scale. arXiv preprint arXiv:2010.11929  (2020)

\bibitem{geiger2015kitti}
Geiger, A., Lenz, P., Stiller, C., Urtasun, R.: The kitti vision benchmark suite. URL http://www. cvlibs. net/datasets/kitti  \textbf{2}(1--5) (2015)

\bibitem{geiger2012we}
Geiger, A., Lenz, P., Urtasun, R.: Are we ready for autonomous driving? the kitti vision benchmark suite. In: Proceedings of the IEEE/CVF Conference on Computer Vision and Pattern Recognition. pp. 3354--3361. IEEE (2012)

\bibitem{gu2021interpreting}
Gu, J., Dong, C.: Interpreting super-resolution networks with local attribution maps. In: Proceedings of the IEEE/CVF Conference on Computer Vision and Pattern Recognition. pp. 9199--9208 (2021)

\bibitem{hendrycks2016gaussian}
Hendrycks, D., Gimpel, K.: Gaussian error linear units (gelus). arXiv preprint arXiv:1606.08415  (2016)

\bibitem{houlsby2019parameter}
Houlsby, N., Giurgiu, A., Jastrzebski, S., Morrone, B., De~Laroussilhe, Q., Gesmundo, A., Attariyan, M., Gelly, S.: Parameter-efficient transfer learning for nlp. In: International Conference on Machine Learning. pp. 2790--2799. PMLR (2019)

\bibitem{hu2021lora}
Hu, E.J., Shen, Y., Wallis, P., Allen-Zhu, Z., Li, Y., Wang, S., Wang, L., Chen, W.: Lora: Low-rank adaptation of large language models. arXiv preprint arXiv:2106.09685  (2021)

\bibitem{jeon2018enhancing}
Jeon, D.S., Baek, S.H., Choi, I., Kim, M.H.: Enhancing the spatial resolution of stereo images using a parallax prior. In: Proceedings of the IEEE conference on computer vision and pattern recognition. pp. 1721--1730 (2018)

\bibitem{jia2022visual}
Jia, M., Tang, L., Chen, B.C., Cardie, C., Belongie, S., Hariharan, B., Lim, S.N.: Visual prompt tuning. In: European Conference on Computer Vision. pp. 709--727. Springer (2022)

\bibitem{jin2022swinipassr}
Jin, K., Wei, Z., Yang, A., Guo, S., Gao, M., Zhou, X., Guo, G.: Swinipassr: Swin transformer based parallax attention network for stereo image super-resolution. In: Proceedings of the IEEE/CVF Conference on Computer Vision and Pattern Recognition. pp. 920--929 (2022)

\bibitem{kim2016accurate}
Kim, J., Lee, J.K., Lee, K.M.: Accurate image super-resolution using very deep convolutional networks. In: Proceedings of the IEEE/CVF Conference on Computer Vision and Pattern Recognition. pp. 1646--1654 (2016)

\bibitem{lei2021deep}
Lei, J., Zhang, Z., Fan, X., Yang, B., Li, X., Chen, Y., Huang, Q.: Deep stereoscopic image super-resolution via interaction module. IEEE Transactions on Circuits and Systems for Video Technology  \textbf{31}(8),  3051--3061 (2021). \doi{10.1109/TCSVT.2020.3037068}

\bibitem{li2021efficient}
Li, W., Lu, X., Qian, S., Lu, J., Zhang, X., Jia, J.: On efficient transformer-based image pre-training for low-level vision. arXiv preprint arXiv:2112.10175  (2021)

\bibitem{liang2021swinir}
Liang, J., Cao, J., Sun, G., Zhang, K., Van~Gool, L., Timofte, R.: Swinir: Image restoration using swin transformer. In: Proceedings of the IEEE/CVF Conference on Computer Vision and Pattern Recognition. pp. 1833--1844 (2021)

\bibitem{lim2017enhanced}
Lim, B., Son, S., Kim, H., Nah, S., Mu~Lee, K.: Enhanced deep residual networks for single image super-resolution. In: Proceedings of the IEEE Conference on Computer Vision and Pattern Recognition Workshops. pp. 136--144 (2017)

\bibitem{lin2023steformer}
Lin, J., Yin, L., Wang, Y.: Steformer: Efficient stereo image super-resolution with transformer. IEEE Transactions on Multimedia pp. 1--13 (2023)

\bibitem{liu2021swin}
Liu, Z., Lin, Y., Cao, Y., Hu, H., Wei, Y., Zhang, Z., Lin, S., Guo, B.: Swin transformer: Hierarchical vision transformer using shifted windows. In: Proceedings of the IEEE/CVF International Conference on Computer Vision. pp. 10012--10022 (2021)

\bibitem{mathai2022lymph}
Mathai, T.S., Lee, S., Elton, D.C., Shen, T.C., Peng, Y., Lu, Z., Summers, R.M.: Lymph node detection in t2 mri with transformers. In: Medical Imaging 2022: Computer-Aided Diagnosis. vol. 12033, pp. 855--859. SPIE (2022)

\bibitem{oquab2023dinov2}
Oquab, M., Darcet, T., Moutakanni, T., Vo, H., Szafraniec, M., Khalidov, V., Fernandez, P., Haziza, D., Massa, F., El-Nouby, A., et~al.: Dinov2: Learning robust visual features without supervision. arXiv preprint arXiv:2304.07193  (2023)

\bibitem{pfeiffer2020adapterfusion}
Pfeiffer, J., Kamath, A., R{\"u}ckl{\'e}, A., Cho, K., Gurevych, I.: Adapterfusion: Non-destructive task composition for transfer learning. arXiv preprint arXiv:2005.00247  (2020)

\bibitem{radford2021learning}
Radford, A., Kim, J.W., Hallacy, C., Ramesh, A., Goh, G., Agarwal, S., Sastry, G., Askell, A., Mishkin, P., Clark, J., et~al.: Learning transferable visual models from natural language supervision. In: International Conference on Machine Learning. pp. 8748--8763. PMLR (2021)

\bibitem{radford2019language}
Radford, A., Wu, J., Child, R., Luan, D., Amodei, D., Sutskever, I., et~al.: Language models are unsupervised multitask learners. OpenAI blog  \textbf{1}(8), ~1--9 (2019)

\bibitem{ruckle2020adapterdrop}
R{\"u}ckl{\'e}, A., Geigle, G., Glockner, M., Beck, T., Pfeiffer, J., Reimers, N., Gurevych, I.: Adapterdrop: On the efficiency of adapters in transformers. arXiv preprint arXiv:2010.11918  (2020)

\bibitem{scharstein2014high}
Scharstein, D., Hirschm{\"u}ller, H., Kitajima, Y., Krathwohl, G., Ne{\v{s}}i{\'c}, N., Wang, X., Westling, P.: High-resolution stereo datasets with subpixel-accurate ground truth. In: Pattern Recognition: 36th German Conference, GCPR 2014, M{\"u}nster, Germany, September 2-5, 2014, Proceedings 36. pp. 31--42. Springer (2014)

\bibitem{vaswani2017attention}
Vaswani, A., Shazeer, N., Parmar, N., Uszkoreit, J., Jones, L., Gomez, A.N., Kaiser, {\L}., Polosukhin, I.: Attention is all you need. Advances in Neural Information Processing Systems  \textbf{30},  1--11 (2017)

\bibitem{wang2021boundary}
Wang, J., Wei, L., Wang, L., Zhou, Q., Zhu, L., Qin, J.: Boundary-aware transformers for skin lesion segmentation. In: Medical Image Computing and Computer Assisted Intervention--MICCAI 2021: 24th International Conference, Strasbourg, France, September 27--October 1, 2021, Proceedings, Part I 24. pp. 206--216. Springer (2021)

\bibitem{wang2019learning}
Wang, L., Wang, Y., Liang, Z., Lin, Z., Yang, J., An, W., Guo, Y.: Learning parallax attention for stereo image super-resolution. In: Proceedings of the IEEE/CVF Conference on Computer Vision and Pattern Recognition. pp. 12250--12259 (2019)

\bibitem{wang2019flickr1024}
Wang, Y., Wang, L., Yang, J., An, W., Guo, Y.: Flickr1024: A large-scale dataset for stereo image super-resolution. In: Proceedings of the IEEE/CVF International Conference on Computer Vision Workshops. pp.~1--6 (2019)

\bibitem{wang2021symmetric}
Wang, Y., Ying, X., Wang, L., Yang, J., An, W., Guo, Y.: Symmetric parallax attention for stereo image super-resolution. In: Proceedings of the IEEE/CVF Conference on Computer Vision and Pattern Recognition. pp. 766--775 (2021)

\bibitem{yang2023aim}
Yang, T., Zhu, Y., Xie, Y., Zhang, A., Chen, C., Li, M.: Aim: Adapting image models for efficient video action recognition. arXiv preprint arXiv:2302.03024  (2023)

\bibitem{ying2020stereo}
Ying, X., Wang, Y., Wang, L., Sheng, W., An, W., Guo, Y.: A stereo attention module for stereo image super-resolution. IEEE Signal Processing Letters  \textbf{27},  496--500 (2020)

\bibitem{Ying2020astere}
Ying, X., Wang, Y., Wang, L., Sheng, W., An, W., Guo, Y.: A stereo attention module for stereo image super-resolution. IEEE Signal Processing Letters  \textbf{27},  496--500 (2020). \doi{10.1109/LSP.2020.2973813}

\bibitem{zaken2021bitfit}
Zaken, E.B., Ravfogel, S., Goldberg, Y.: Bitfit: Simple parameter-efficient fine-tuning for transformer-based masked language-models. arXiv preprint arXiv:2106.10199  (2021)

\bibitem{zhang2023adaptive}
Zhang, Q., Chen, M., Bukharin, A., He, P., Cheng, Y., Chen, W., Zhao, T.: Adaptive budget allocation for parameter-efficient fine-tuning. arXiv preprint arXiv:2303.10512  (2023)

\bibitem{zhang2018image}
Zhang, Y., Li, K., Li, K., Wang, L., Zhong, B., Fu, Y.: Image super-resolution using very deep residual channel attention networks. In: Proceedings of the European conference on computer vision. pp. 286--301 (2018)

\bibitem{zhang2018residual}
Zhang, Y., Tian, Y., Kong, Y., Zhong, B., Fu, Y.: Residual dense network for image super-resolution. In: Proceedings of the IEEE/CVF Conference on Computer Vision and Pattern Recognition. pp. 2472--2481 (2018)

\bibitem{zhou2022learning}
Zhou, K., Yang, J., Loy, C.C., Liu, Z.: Learning to prompt for vision-language models. International Journal of Computer Vision  \textbf{130}(9),  2337--2348 (2022)

\bibitem{zhou2023stereo}
Zhou, Y., Xue, Y., Deng, W., Nie, R., Zhang, J., Pu, J., Gao, Q., Lan, J., Tong, T.: Stereo cross global learnable attention module for stereo image super-resolution. In: Proceedings of the IEEE/CVF Conference on Computer Vision and Pattern Recognition. pp. 1416--1425 (2023)

\end{thebibliography}
\end{document}